\documentclass[journal]{IEEEtran}
\usepackage[T1]{fontenc}
\usepackage[latin1]{inputenc}
\usepackage[english]{babel}
\usepackage{siunitx}
\usepackage{amsmath}
\usepackage{mathtools}
\usepackage{amsfonts}
\usepackage{float}
\usepackage{caption}
\usepackage{subcaption}
\usepackage{hyperref}

\begin{document}
\title{Optical aberrations in autonomous driving: Physics-informed parameterized temperature scaling for neural network uncertainty calibration}

\author{Dominik~Werner~Wolf,~\IEEEmembership{Member,~IEEE,}
        Alexander~Braun,~\IEEEmembership{Non-Member,}
        and~Markus~Ulrich,~\IEEEmembership{Non-Member}
\thanks{D.~W.~Wolf is with the Machine Vision Metrology Group of the Karlsruhe Institute of Technology~(KIT) and conducts research for the Volkswagen Group. E-mail:~\url{dominik.werner.wolf@volkswagen.de}.}%
\thanks{M.~Ulrich is with the Karlsruhe Institute of Technology (KIT).}%
\thanks{A.~Braun is with the University of Applied Sciences Duesseldorf (HSD).}}%

\markboth{IEEE Transactions on Intelligent Transportation Systems,~Vol.~XX, No.~X, April~2025}%
{Shell \MakeLowercase{\textit{et al.}}: Bare Demo of IEEEtran.cls for IEEE Journals}

\maketitle

\begin{abstract}
\textit{``A trustworthy representation of uncertainty is desirable and should be considered as a key feature of any machine learning method''}\;\cite{Uncertainty_decomposition}. This conclusion of Huellermeier et al. underpins the importance of calibrated uncertainties. Since AI-based algorithms are heavily impacted by dataset shifts, the automotive industry needs to safeguard its system against all possible contingencies. One important but often neglected dataset shift is caused by optical aberrations induced by the windshield. For the verification of the perception system performance, requirements on the AI performance need to be translated into optical metrics by a bijective mapping. Given this bijective mapping it is evident that the optical system characteristics add additional information about the magnitude of the dataset shift. As a consequence, we propose to incorporate a physical inductive bias into the neural network calibration architecture to enhance the robustness and the trustworthiness of the AI target application, which we demonstrate by using a semantic segmentation task as an example. By utilizing the Zernike coefficient vector of the optical system as a physical prior we can significantly reduce the mean expected calibration error in case of optical aberrations. As a result, we pave the way for a trustworthy uncertainty representation and for a holistic verification strategy of the perception chain.
\end{abstract}

\begin{IEEEkeywords}
Neural Network Calibration, Predictive Uncertainty, Dataset shifts, Wavefront aberrations, Autonomous driving.
\end{IEEEkeywords}

\section{Introduction}
Autonomously driving cars perceive the environment through different sensors, e.g., wide-angle cameras, telephoto cameras~\cite{Mobileye, Tesla, Nvidia}, Light Detection and Ranging~(LiDAR) sensors etc. Typically, the measured sensor signals serve as the input to a neural network, which is supposed to predict the actions required (e.g., acceleration, steering angle etc.) to reach the next state. In the development phase, the neural network is trained on a training dataset that is typically captured by a small fleet of test mules. If everything goes well with the architectural design and the neural network demonstrates sufficient accuracy and generalizability, then the autonomous driving functionality might be considered as operational. As the conceptual phase is taken to serial production, the real-world performance of the AI-based driving function might differ by a huge margin from what has been observed during development. A prominent contributor to this phenomenon is the perception chain of the telephoto camera (i.e., a camera with a telephoto lens). Telephoto cameras are distinguished by a long focal length, which results in a high pixel resolution per field-angle. As a consequence, the target application of telephoto cameras is object detection and classification for far-field objects, especially important on highways with high driving speeds. Unfortunately, this benefit also comes with a downside, namely an increased sensitivity for optical aberrations. Every car has a windshield, and every windshield has its unique aberration pattern. This has not been an issue for standard automotive cameras because the width of the blurring kernel induced by the windshield was always smaller than a pixel-pitch of the Complementary Metal-Oxide-Semiconductor~(CMOS) sensor. With the use of telephoto cameras, this does not hold true anymore and the images captured might by heavily impacted by the windshield in terms of sharpness. This gives rise to a shift in image quality between the test mule recordings used for training and the car-by-car perception chain.

Dataset shifts are of major concern for the homologation of safety-critical autonomous driving functions. Dataset shifts are generally given if the network infers information from an instance, which does not share the same underlying probability density function~(PDF) as the training dataset distribution. Consequently, the neural network utilizes the learned functional relationship between input and output for extrapolating into a different domain. This gives rise to a performance drop of the model. The performance is not only affected in terms of the target key performance indicator~(KPI) but also the corresponding uncertainty estimation might become biased~\cite{Wolf_ICCV2023}.

According to information theory, the total predictive uncertainty for a classification problem is given by the Shannon entropy~\cite{Uncertainty_decomposition}. From a metrological perspective, if the model is properly calibrated, the uncertainty measure is expected to align with the observed error rate in the network's predictions. Equivalently speaking, a perfectly calibrated network is given if the confidence estimate is congruent to the measured prediction accuracy. For assessing the calibration performance, there exists an entire zoo of measures, e.g., the Uncertainty Calibration Error~(UCE)~\cite{mECE_mUCE}, the Area Under the Sparsification Error curve~(AUSE$_{S}$)~\cite{dreissig2023calibration} utilizing the Shannon entropy~\cite{Shannon} for sorting, the Expected Calibration Error~(ECE)~\cite{mECE_mUCE} or the AUSE$_{V}$ utilizing the variation ratio~\cite{maag2020time} for sorting. The conceptual differences between point-wise predictive uncertainty calibration estimators~(ECE,~AUSE$_{V}$) and entropy-based calibration measures~(UCE,~AUSE$_{S}$) results in a decoupling of neural network calibration measures~\cite{Wolf_GCPR24}. As a consequence, it is essential to make a physical sound decision on which calibration measure to employ for the neural network under consideration.

In this paper, we focus on semantic segmentation because of the hypothesis that a dataset shift in terms of sharpness will affect a pixel-wise prediction the most. At this point, it is important to underscore that, irrespective of the specific AI task selected, this methodology exhibits encouraging potential for effective generalization across a broad spectrum of tasks. Our semantic segmentation Convolutional Neural Network~(CNN) will employ a negative log-likelihood loss, which makes it favorable to rely on point-wise predictive uncertainty calibration estimators because it matches the nature of the ground truth label distribution, which allocates all statistical mass to the ground truth class and analytically resembles the Kronecker delta function. As a consequence, the expected Shannon information~\cite{Shannon} is given by the negative logarithm of the probability score predicted by the CNN for the ground truth class and the cross-entropy is minimized by maximizing the prediction confidence for the ground truth class during training. As a result, the expected negative log-likelihood loss, aka. cross-entropy, is invariant under differences in the probability mass allocation over the remaining wrong classes. The same does not hold true for the Shannon entropy~\cite{Shannon}, which is the standard measure for the total predictive uncertainty according to information theory~\cite{Uncertainty_decomposition}. This gives rise to a degree of freedom in the uncertainty evaluation or to put it differently, the entropy-based uncertainties for two independent instances might differ even though the model confidence predictions are equivalent~\cite{Wolf_GCPR24}. We avoid this by employing the variation ratio~\cite{maag2020time} as a point-wise measure for the prediction uncertainty and the ECE as a point-wise calibration measure.

Calibrated uncertainties are an essential requirement for a physically sound sensor fusion process and for system monitoring. On the one hand, fusing feature attributes should incorporate the associated embedding uncertainties in order to achieve the most reliable latent space representation. On the other hand, in order to safeguard autonomous systems, the prediction uncertainties need to be tracked. If the uncertainty is low, and hence the situation is identified with sufficient confidence, a reliable decision can be made. If the confidence is insufficiently low then an independent secondary system must contribute additional information for the decision-making process or the system has to fall back into a safe state mode automatically. Consequently, the trustworthiness of the uncertainty estimates, quantified by the ECE, is decisive for the reliability of autonomous driving systems. However, if a dataset shift is induced, e.g. by optical aberrations of the windshield, the calibration of the network confidences - and uncertainties vice versa - breaks down and the network becomes increasingly overconfident~\cite{Wolf_ICCV2023}.

To tackle this task, we present a novel neural network architecture, which extends the state-of-the-art Parameterized Temperature Scaling~(PTS)~\cite{PTS} approach by incorporating a physical inductive bias~\cite{PINNs_CV}. We demonstrate the benefits of integrating physical priors to PTS, which we will refer to as Physics Informed Parameterized Temperature Scaling~(PIPTS), by comparing the results to the state-of-the-art PTS method and to the standard Temperature Scaling~(TS)~\cite{TS} technique. The physical prior consists of the predicted Zernike coefficient vector of the optical system and is intended to enhance the resilience of the autonomous driving perception pipeline against optical perturbations.

In our work, we combine advanced methods from two different domains, machine learning and optics. We appreciate that practitioners from each field might find the other domain challenging, but refrain from too long theoretical introductions and instead refer to the literature.

\section{Related work}
Fundamentally, the temperature determines the sensitivity of the entropy w.r.t.\ changes in the internal energy from a physical perspective~\cite{States_of_matter_Goodstein} and w.r.t.\ changes in the logits from an AI point of view~\cite{Shannon}.

In physics, low temperature means atoms move slowly and occupy minimal-energy states. As temperature rises, atoms gain energy, which makes higher-energy states accessible and the state variability increases.

In AI language models, temperature controls the variability in word predictions~\cite{Calibrating_Language_Models}. The model predicts a likelihood for each word within the vocabulary and the subsequent word is chosen randomly according to the predicted probability mass function. Calibrating the model with a temperature of zero leads to deterministic behavior and complete repeatability. Increasing the temperature allows words with lower likelihoods to still have a chance, creating more diverse results. This process mirrors the Boltzmann distribution in physics, where states are sampled based on energy levels. In AI, the Softmax function performs a similar task, treating model logits as negative energies. Higher temperatures broaden the probability distribution, making all outcomes more equal, while lower temperatures sharpen the probability distribution.

Finding the optimal temperature is the key for establishing model trustworthiness. The straightforward way to determine the temperature is given by minimizing the negative log-likelihood~\cite{TS}. Unfortunately, this standard Temperature Scaling~(TS) methodology is highly limited in terms of the model information capacity~(one degree of freedom).

As an extension, Ensemble Temperature Scaling~(ETS)~\cite{ETS} computes an weighted average over three different calibration maps, the TS calibrator with adjustable temperature~$T$, TS with $T = 1$~(identity mapping) and TS with $T = \infty$~(uniform mapping). Hence, ETS has four degrees of freedom.

In order to further increase the information capacity of the calibration method, Parameterized Temperature Scaling~(PTS)~\cite{PTS} was proposed. PTS leverages a neural network to predict an instance-wise temperature based on the corresponding logit tensor, while preserving model accuracy.

A similar methodology is Sample-Dependent Adaptive Temperature Scaling~\cite{Sample_dependent_ATS}, which also predicts an instance-wise temperature. In contrast to PTS, the approach leverages the latent space representation of a Variational Autoencoder~(VAE)~\cite{Variational_Autoencoders} as the input for a post-hoc Multi-Layer Perceptron~(MLP) for predicting an instance-wise temperature. The benefit of using the VAE's latent space embeddings instead of the logit tensor for the post-hoc MLP lies in the effectiveness of the VAE to cluster the predictions based on their calibration quality, which improves the calibration performance of the MLP under distribution shifts~\cite{Sample_dependent_ATS}.

Most recently, Adaptive Temperature Scaling~(ATS)~\cite{ATS} has been proposed as a calibration technique, which enhances the reliability against out-of-distribution samples without the need of training a post-hoc MLP calibrator. The core idea of ATS lies in computing an instance-wise temperature based on the intermediate layer activations of the baseline neural network. After training, the Cumulative Distribution Function~(CDF) of the mean activation for each layer is computed across the entire training dataset. During inference, the layer-wise mean activations are compared to the precomputed CDFs to calculate layer-wise p-values~\cite{p_values, Oxford_Dictionary_of_Statistics}, which are mapped to an instance-wise temperature. In addition to an enhanced calibration performance, low temperatures indicate in-distribution samples, while high temperatures suggest out-of-distribution inputs~\cite{ATS}.

For all accuracy preserving calibration methods mentioned so far, it is required to assume an uncertainty estimator. In mathematical terms, calibration quality metrics measure the bias of a predictive uncertainty estimator. Consequently, neural network calibration and predictive uncertainty estimation are two distinct concepts. A perfect predictive uncertainty estimator is unbiased such that the calibration temperature is equal to one. There are several different methods to estimate the predictive uncertainty for semantic segmentation~\cite{survey_uncertainty_DNN}. A very powerful approach are Deep Ensembles~\cite{Deep_ensembles} that use several neural networks of the same architecture in parallel but with different initializations to retrieve various subsamples from the posterior distribution. The mean of this subsample is outputted as the model's prediction and the standard deviation of the mean serves as the predictive uncertainty estimator. A major downside of this approach is the computational overhead generated during training and inference by the utilization of several models that sample the hypothesis space~\cite{Uncertainty_decomposition}. This problem has been addressed recently in the work of Landgraf et al. on Deep Uncertainty Distillation using Ensembles for Semantic Segmentation~(DUDES)~\cite{DUDES}. They propose a student-teacher distillation framework where the Deep Ensemble model, referred to as the teacher, is used to guide a less complex model, known as the student, in estimating the ensemble-based predictive uncertainties. This significantly reduces the inference time because only one forward pass is required and the information about the posterior distribution is distilled into the student neural network, wherefore the calibration quality of the predictive uncertainty estimate is maintained. This concept has been further extended by the student-teacher distillation framework for efficient multi-task uncertainties, referred to as EMUFormer~\cite{EMUFormer}. They employ a Deep Ensemble of transformer-based multi-task networks for semantic segmentation and monocular depth estimation (termed as SegDepthFormer) to evaluate the predictive uncertainty. The backbone of this methodology is the idea of enhancing the generalization capabilities of a neural network by multi-task learning. With the SegDepthFormer architecture they demonstrated that this idea can be transformed to the network calibration and the results indicate less biased predictive uncertainty estimates in terms of the mECE if multi-task learning is employed. Nevertheless, we will select the variation ratio~\cite{maag2020time} as a measure for the predictive uncertainty in this work because of the non-existent computational overhead and it's simplicity, which mainly explains why the variation ratio is widely adapted.

In our use case we know what drives the distribution shift, namely optical aberrations within the perception chain, e.g., the windshield or diverse weather phenomena. As a consequence, it seems natural to incorporate this prior knowledge into the calibration process. In order to do so, the optical aberrations need to be estimated online alongside the target application. The most fundamental way to characterize optical aberrations is given by quantifying the optical path difference map in terms of the Zernike coefficients~\cite{Zernike_I, Zernike_II, Zernike_III}. Utilizing a neural network for predicting the Zernike coefficient vector is a well established approach in astronomy~\cite{Astronomy_Zernike_idea, Astronomy_Zernike_ResNet} as a way to replace the need for on the fly Shack-Hartmann measurements. The Very Large Telescope~(VLT) of the European Southern Observatory~(ESO) uses the information about the Zernike coefficients to perform an online correction of the wavefront aberrations induced by the atmosphere~\cite{adaptive_optics_EOS}. This is realized by adjusting deformable mirrors, which is a technique from adaptive optics~\cite{adaptive_optics}.

The work of Jaiswal et al. on physics-driven turbulence image restoration with stochastic refinement~\cite{Physics_Driven_Restoration} utilizes the Zernike coefficients to parameterize a physics-based turbulence simulator. By coupling the vision transformer-based~\cite{vision_transformer} image restoration network with the Fourier-optical aberration model during training, they are able to effectively disentangle the stochastic degradation caused by atmospheric turbulence from the underlying image. This enhances the generalizability of the image restoration network across real-world datasets with varying turbulence strength~\cite{Physics_Driven_Restoration}. As a consequence, by incorporating a physical inductive bias to the transformer architecture they effectively reduce the sensitivity of their target application on dataset shifts induced by turbulence-driven optical aberrations.

In our work, we want to seize the idea of coupling physical priors with the baseline neural network architecture in order to enhance the calibration robustness.

\section{Theoretical essentials}
In this section, we want to briefly introduce the relevant concepts that are used within this work. First, the optical merit functions of interest will be specified. Subsequently, we will define the relevant measures from the AI world. With this framework parameterization we will investigate the dependency of the neural network performance for semantic segmentation on the optical quality of the perception chain. The non-linear correlation between those KPIs can be quantified by the Chatterjee's rank correlation measure~\cite{Chatterjee_correlation, Dette-Siburg-Stoimenov_correlation}. We will shortly address the theoretical foundations of the Chatterjee's rank correlation measure as it will be required in order to select the most suitable optical metric.

\subsection{Optical merit functions}
Within this work, we are studying three different optical metrics, which have been used in previous work by Wolf~et~al.~\cite{Wolf_ICCV2023} for the sensitivity analysis of AI-based algorithms for autonomous driving on optical wavefront aberrations induced by the windshield. These optical metrics can be evaluated if the wavefront aberration map is known a priori~\cite{Wolf_ICCV2023, Wolf_ITSC2023}. Generally, the wavefront aberration map~$W$ is decomposed into the orthogonal Zernike polynomial basis~$Z_{n}$~\cite{Zernike_I, Zernike_II, Zernike_III} parameterized by the corresponding Zernike coefficients~$\alpha_{n}$:
\begin{equation}
    W(\rho_{r},\;\phi_{a}) = \sum \limits_{n=0}^{\infty} \alpha_{n} Z_{n}(\rho_{r},\;\phi_{a}) \;\;\;\text{,}\;\;\; \alpha_{n} := \left<W,\;Z_{n}\right>\;.
\label{eq:Zernike_decomposition}
\end{equation}
From the optical path difference distribution across the aperture surface, parameterized by the normalized radial coordinate~$\rho_{r}$ and the azimuth angle~$\phi_{a}$, the Point Spread Function~(PSF) can be calculated by applying Fourier optical principles~\cite{Fourier_optics}. In a nutshell, the PSF is the impulse response function or Green's function of an optical system, which entirely determines the behavior of a Linear and Time-Invariant~(LTI) system~\cite{Control_Systems}.

The Fourier transform of the real-valued, incoherent PSF is known as the Optical Transfer Function~(OTF). The OTF is generally complex-valued if the PSF is non-symmetric w.r.t.\ the optical axis. If the PSF is viewed as a scaled probability density function of the light distribution in the observer plane, then the PSF can be entirely characterized by its statistical moments~\cite{Green_as_PDF, Wolf_ICCV2023} and the OTF serves as the corresponding characteristic function. Consequently, the k-th order derivative of the OTF at zero spatial frequency entirely determines the k-th moment of the light distribution, e.g., gray values centroid~(k=1), intensity variance~(k=2) etc.

The automotive industry is currently trying to grasp the importance of the OTF as an optical quality indicator function. Unfortunately, current attempts to map the OTF to a real-valued, scalar metric lead to insufficient optical KPIs as the Modulation Transfer Function~(MTF) at half-Nyquist frequency~\cite{Wolf_ITSC2023, MTF_requirement}. Analytically, the MTF is defined as the real part of the OTF~\cite{Fourier_optics}.

As an extension to mapping the OTF to a single spatial frequency value of the MTF, the Strehl ratio~\cite{Fourier_optics} has been proposed as an alternative measure to incorporate information about the entire spectrum into the mapping process. The Strehl ratio is defined as the spectral integral of the MTF in relation to the diffraction limited MTF area~\cite{Fourier_optics}.

An attempt to distill even more information into the mapping process of the OTF was made by Wolf~et~al.~\cite{Wolf_ICCV2023}. They proposed the Optical Informative Gain~(OIG) as the normalized spectral integral of the squared MTF function. This minor adjustment of the definition of the Strehl ratio is theoretically beneficial because the Strehl ratio exclusively captures information about the PSF at the optical axis. Hence, the captured information about the PSF, which entirely characterizes the optical system, is very limited and higher-order statistical moments are not accounted for at all. The OIG alleviates this situation by exploiting the Plancherel theorem~\cite{Plancherel_theorem} to retrieve information about the energy, which can be spatially discriminated in relation to the diffraction-limited case~\cite{Wolf_ICCV2023}.

\subsection{Neural network KPIs}
We will study the calibration quality of the pixel-wise confidences predicted by a CNN-based decoder head for semantic segmentation. The performance of the multi-class semantic segmentation task is evaluated by the mean Intersection over Union~(mIoU)~\cite{Survey_paper}:
\begin{equation}
    \mathrm{mIoU} := \dfrac{1}{N_{c}}\sum\limits_{i=0}^{N_{c}} \resizebox{0.125\linewidth}{!}{$ \dfrac{\mathrm{G}_{i} \cap \mathrm{P}_{i}}{\mathrm{G}_{i} \cup \mathrm{P}_{i}} $} \; \hat{=} \; \dfrac{1}{N_{c}}\sum\limits_{i=0}^{N_{c}} \resizebox{0.275\linewidth}{!}{$ \dfrac{\mathrm{TP}_{i}}{\mathrm{TP}_{i} + \mathrm{FP}_{i} + \mathrm{FN}_{i}} $}\;.
\label{eq:mIoU}
\end{equation}
Here, $N_{c}$ is the number of classes, $\mathrm{G}_{i}$ represents the set of ground truth labels and $\mathrm{P}_{i}$ indicates the set of predictions for class $i$. In detail, the number of class-wise true positive predictions ($\mathrm{TP}_{i}$) is normalized by the total number of predictions within the cross-section domain composed by $\mathrm{TP}_{i}$, the class-wise false positive predictions $\mathrm{FP}_{i}$, and the class-wise false negative predictions $\mathrm{FN}_{i}$.

For assessing the calibration performance we will employ the mean ECE~(mECE) as a point-wise calibration measure over all $N_{c}$ classes. The mECE, introduced by Naeini~et~al.~\cite{mECE_mUCE}, is the weighted and binned average of the absolute difference between the model accuracy~($\mathrm{acc}$) and the confidence~($\mathrm{conf}$). We will employ the softmax likelihood as a confidence estimator in alignment with the variation ratio~\cite{maag2020time}. Mathematically, the mECE is given by:
\begin{equation}
    \mathrm{mECE} := \sum_{m=1}^{N_{b}} \frac{\left|B_m\right|}{N_{c}}\left|\operatorname{acc}\left(B_m\right) - \operatorname{conf}\left(B_m\right)\right|\;,
\label{eq:ece}
\end{equation}
where the pixel-wise predictions are binned according to their confidence score into~$N_{b}$ bins~$B_m$ of equal width in the range~$[0,1]$.

Finally, we define the Area Under the Reliability Error Curve~(AUREC) as an additional point-wise calibration measure for classification, inspired by the UCS calibration metric for regression tasks~\cite{UCS_Kira}. The AUREC is equivalent to the mECE except that the weighting factor is set to one in order to artificially magnify the impact of predictions that fall into low confidence bins, which typically show a low cardinality~\cite{Wolf_GCPR24}.

\subsection{Non-linear correlation measure}
\begin{figure}[b]
   \centering
   \includegraphics[width=1.\linewidth]{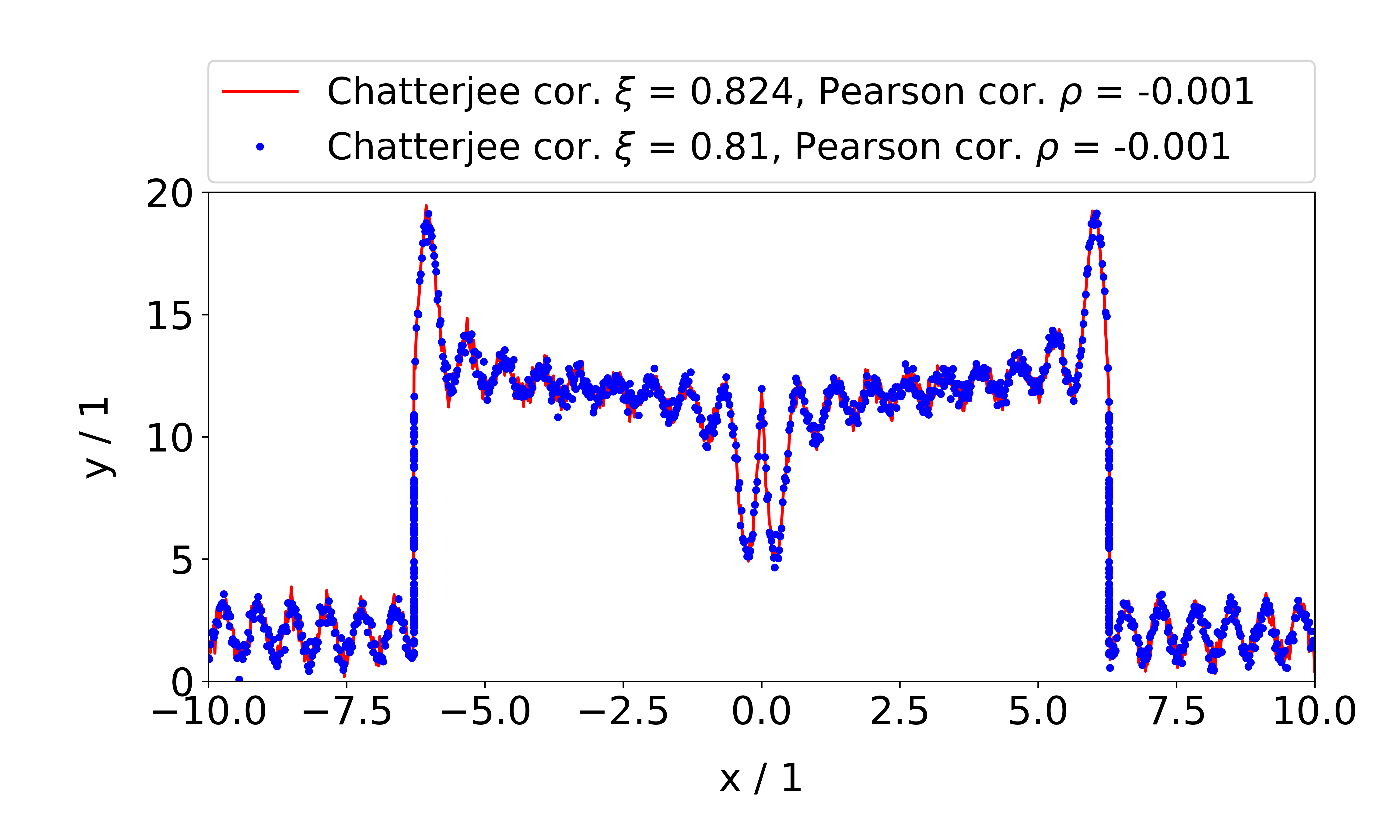}
   \caption{The Chatterjee's rank correlation measure~$\xi$ is compared to the Pearson correlation coefficient~$\rho$ for the test function presented in Equation~(\ref{eq:test_function}). The red curve indicates the functional relationship with Gaussian noise applied to it. Furthermore, subsamples are added randomly at the discontinuity~($x = \pm 2\pi$) to demonstrate the sensitivity of~$\xi$ on the unexplained variance contribution, depicted by the blue dots.}
   \label{fig:Chatterjee_correlation_test}
\end{figure}

In order to quantify the correlation between two input signals, suppose $x$ and $y$, the Pearson correlation coefficient~$\rho$ is typically employed. Unfortunately, the Pearson correlation coefficient is restricted to linear relationships. In order to evaluate the non-linear correlation between the two signals -- in our case 'optical quality' and 'AI performance' -- alternative metrics are required. A very fundamental definition for non-linear correlation measures is given by the Dette-Siburg-Stoimenov's rank correlation metric~$\left<\xi_{n}\right>$~\cite{Dette-Siburg-Stoimenov_correlation} defined as:
\begin{align}
\begin{split}
    \left<\xi_{n}\right> &\coloneqq \dfrac{\left< \mathrm{VAR}_{x} \left[ \mathrm{E}_{y} \left[ \mathbf{1}_{\left\{ y \geq t \right\}}(y) \; | \; x \right] \right] \; | \; \mathrm{pdf}_{y}(t) \right>}{\left< \mathrm{VAR}_{y} \left[ \mathbf{1}_{\left\{ y \geq t \right\}}(y) \right] \; | \; \mathrm{pdf}_{y}(t) \right>}\\[5pt]
    \Leftrightarrow \left<\xi_{n}\right> &\;= 1 - \resizebox{0.625\linewidth}{!}{$
    \dfrac{\left< \mathrm{E}_{x} \left[ \mathrm{VAR}_{y} \left[ \mathbf{1}_{\left\{ y \geq t \right\}}(y) \; | \; x \right] \right] \; | \; \mathrm{pdf}_{y}(t) \right>}{\left< \mathrm{VAR}_{y} \left[ \mathbf{1}_{\left\{ y \geq t \right\}}(y) \right] \; | \; \mathrm{pdf}_{y}(t) \right>}
    $}\;.
\end{split}
\label{eq:Dette_Siburg_Stoimenovs_correlation}
\end{align}
The quotient of the second term is given by the expected unexplained variance over the expected total variance. According to the law of variance decomposition~\cite{variance_decomposition}, the total variance~$\mathrm{VAR}\left[y\right]$ is given as the sum of the explained variance~$\mathrm{VAR}\left[ \mathrm{E} \left[ y \; | \; x \right] \right]$ and the unexplained variance~$\mathrm{E}\left[ \mathrm{VAR} \left[ y \; | \; x \right] \right]$. If there is a functional relationship between $x$ and $y$ then the unexplained variance vanishes. The expectation value of the unexplained variance of the indicator function~$\mathbf{1}_{\left\{ y \geq t \right\}}(y)$ over the distribution of $y$ is required in order to scan through all possible ranking thresholds~$t$ according to their likelihood. Hence, the Dette-Siburg-Stoimenov's correlation coefficient is a rank correlation metric.

Equation~(\ref{eq:Dette_Siburg_Stoimenovs_correlation}) is hard to evaluate numerically given a discrete sample. The Chatterjee's rank correlation coefficient~$\xi$~\cite{Chatterjee_correlation} presents an approximation of $\left<\xi_{n}\right>$ that converges to the expectation value as the sample size $n \mapsto \infty$. The Chatterjee's rank correlation coefficient is given by:
\begin{align}
\begin{split}
    \xi_{n} &\coloneqq 1 - \dfrac{n}{2} \cdot \resizebox{0.225\linewidth}{!}{$
    \dfrac{\sum \limits_{i=1}^{n-1} |r_{i+1} - r_{i}|}{\sum \limits_{i=1}^{n} l_{i} \left( n - l_{i} \right)}$} \;\; \resizebox{0.425\linewidth}{!}{$ 
    \text{with:} \;
    \left\{\begin{array}{l}
         r_{i} \coloneqq \sum \limits_{j=1}^{n} \mathbf{1}_{\left\{ y_{j} \leq y_{i}\right\}}\;, \\
         l_{i} \coloneqq \sum \limits_{j=1}^{n} \mathbf{1}_{\left\{ y_{j} \geq y_{i}\right\}}\;.
    \end{array}\right.
    $}
\end{split}
\label{eq:Chatterjees_correlation}
\end{align}
In order to demonstrate the powerfulness of the Chatterjee's rank correlation measure, a toy example case study is presented. Suppose the following test function:
\begin{equation}
    y := \resizebox{0.825\linewidth}{!}{$\left\{ \begin{array}{ll}
        2 - \cos{(10x)} & ,\;x \in \left( -\infty,\;-2\pi \right) \cup \left( 2\pi,\;\infty \right) \\[5pt]
        12 + \sum \limits_{n=1}^{10} \sin{(n x)} & ,\;x \in \left[ -2\pi,\;0 \right) \\[10pt]
        12 - \sum \limits_{n=1}^{10} \sin{(n x)} & ,\;x \in \left[ 0,\;2\pi \right]\;.
    \end{array} \right.$}
\label{eq:test_function}
\end{equation}
Additionally, the function is disturbed by random noise sampled from a Gaussian with zero mean and a standard deviation of $\sigma_{\epsilon} = 0.3$. The corresponding graph is visualized in Figure~\ref{fig:Chatterjee_correlation_test} for~$x \in \left[ -10,\;10 \right]$ sampled uniformly and the corresponding Chatterjee's rank correlation measure amounts to $\xi_{1001} = 0.824$ considering~$n=1001$ samples. If the Pearson correlation coefficient is evaluated in comparison it can be noticed that $\rho$ almost vanishes. This indicates the insufficiency of $\rho$ for non-linear relationships.

\begin{figure}[t]
   \centering
   \includegraphics[width=1.\linewidth]{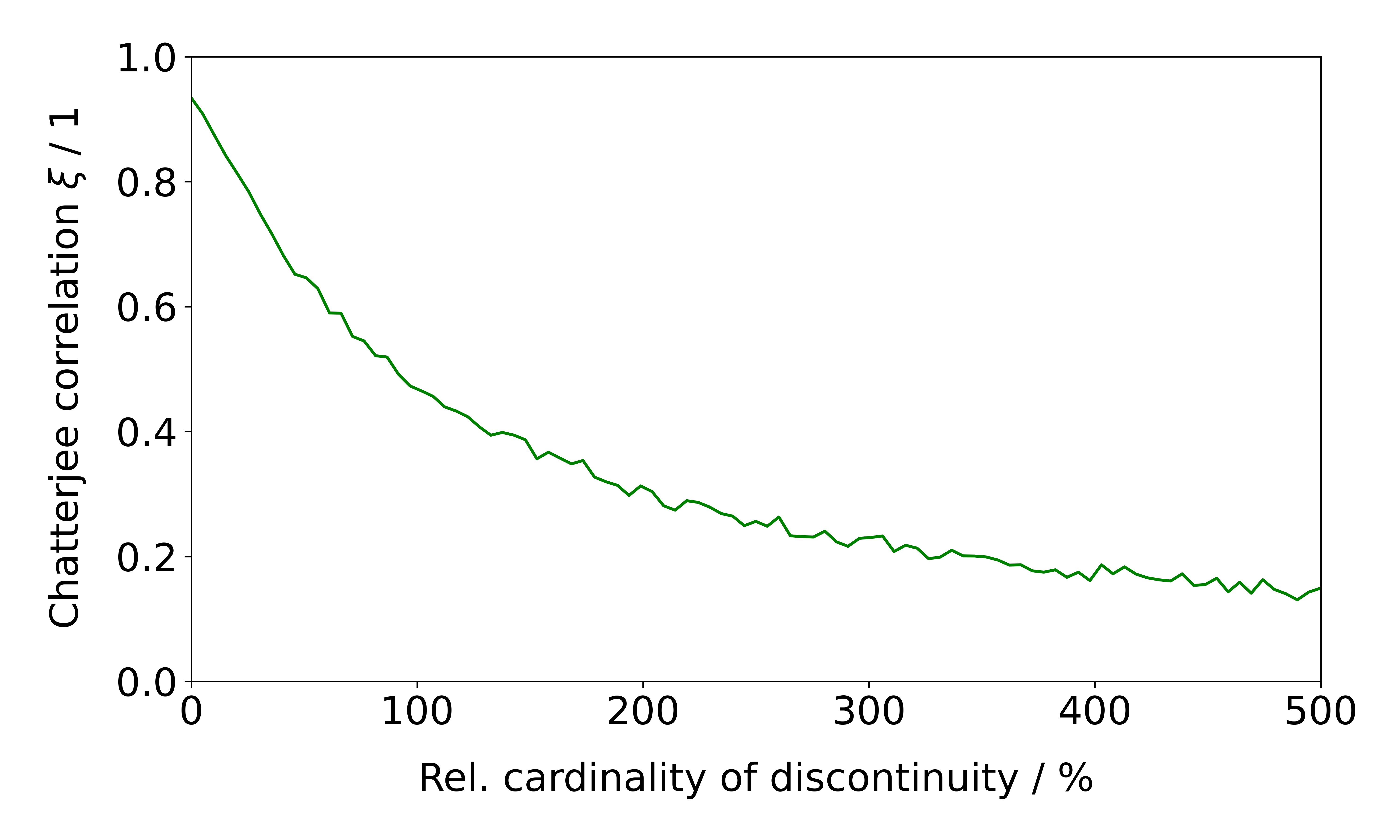}
   \caption{The Chatterjee's rank correlation measure~$\xi$ is shown as a function of the relative cardinality of the subsample inserted at the location of the test function discontinuity at $x = \pm 2\pi$.}
   \label{fig:Chatterjee_correlation_decay}
\end{figure}

Since the Chatterjee's rank correlation measure quantifies the amount of unexplained variance within the sample, it is expected that $\xi_{n}$ reduces if multiple samples are drawn within the discontinuity at~$x = \pm 2\pi$. This is also illustrated in Figure~\ref{fig:Chatterjee_correlation_test}, where $n_{\mathrm{sub}}=100$ subsamples were randomly added within each discontinuity increasing the unexplained variance contribution. The decay of $\xi_{n}$ with increasing cardinality of the inserted subsample at each discontinuity is studied more systematically in Figure~\ref{fig:Chatterjee_correlation_decay}.

\section{PIPTS calibration architecture for semantic segmentation}
This work aims to demonstrate the benefits of incorporating physical priors into the PTS approach for confidence calibration. Since we are concerned with dataset shifts induced by optical aberrations of the windshield, our physical prior consists of the Zernike coefficient vector. In this paper, semantic segmentation is selected as the AI target application due to its intrinsic sensitivity to the impact of the blurring operator, as it is based on pixel-level classifications.

\subsection{Physical priors}
\begin{figure*}[!t]
   \centering
   \includegraphics[width=1.\linewidth]{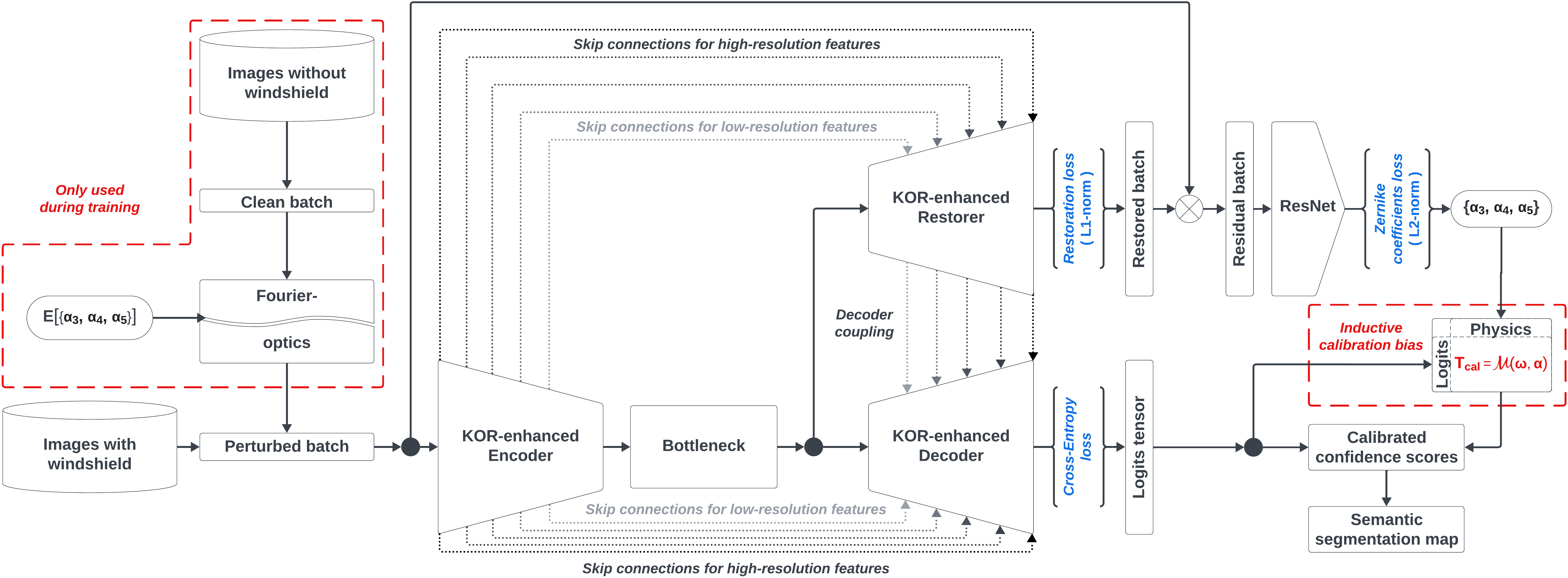}
   \caption{The layout of the multi-task network for semantic segmentation and for predicting the effective Zernike coefficients of the optical system is shown. The multi-task network builds upon the UNET architecture with two coupled decoder heads and a downstream ResNet encoder for retrieving the Zernike coefficients of the second radial order. Additionally, the Fourier optical degradation model for the data augmentation process and the post-hoc PIPTS calibration network are indicated. The PIPTS calibrator extends the PTS approach by incorporating a physical inductive bias for ensuring the trustworthiness of the baseline multi-task network predictions under optical aberrations.}
   \label{fig:PIPTS_architecture}
\end{figure*}

Generally, optical aberrations can be decomposed into distortion and blurring~\cite{Wolf_ITSC2023, Tilt_vs_Blur}. The tilt operator describes the impact of distortions on the imaging process and is parameterized by the Zernike coefficients of the first radial order. On the other hand, the blurring operator is parameterized by the Zernike coefficients of radial order greater than one. In this study, we will solely focus on the impact of the blurring operator on the imaging process, as a measurement procedure to correct for distortions is already available~\cite{Wolf_Metrologia}. Considering the dominant optical aberrations of windshields, we restrict the Zernike coefficient vector to the coefficients of the second radial order, which includes oblique astigmatism~($\alpha_3$), defocus~($\alpha_4$), and orthogonal astigmatism~($\alpha_5$). To model the effect of these aberrations, we apply a Fourier optical degradation model~\cite{Wolf_ICCV2023, Fourier_optics} for data augmentation. The Zernike coefficients are sampled uniformly in the range~$\alpha_{n} \in [-\lambda, \lambda]$. This does not guarantee that the complexity of the production process is sufficiently reflected, but it serves as a physically sound baseline for our proof of concept study. For a comprehensive modeling of the statistical complexity of the perception chain's manufacturing process, individual part tolerances and installation tolerances need to be accounted for.

\subsection{Architecture of the baseline multi-task network}
In order to solve the target task and to provide the physical prior for the PIPTS calibration network, we propose a CNN-based multi-task architecture with two coupled decoder heads and an additional residual encoder for estimating the Zernike coefficients of the second radial order of the effective wavefront aberration map of the overall optical system consisting of windscreen and ADAS camera lens. The detailed architecture of the baseline multi-task network is illustrated in Figure~\ref{fig:PIPTS_architecture}.

The \textbf{shared encoder} consists of five encoder blocks, each with two convolution layers, two batch normalization layers, and a max pooling layer for downsampling. The number of filters is doubled in each successive encoder block and after the last batch normalization layer, the computational graph is split into two branches to bypass information to the decoder for alleviating the vanishing gradient problem~\cite{ResNet, ResNet_VGP, ResNet_VGP_norm}.

The \textbf{bottleneck} of the UNET, distinguished by the lowest spatial feature resolution, consists of two convolution layers and subsequent batch normalization layers. After the bottleneck, the latent space representation is supposed to provide an embedding of the input information that optimally reflects the degrees of freedom of the underlying problem. Subsequently, the embedding is fed into two decoder heads.

The \textbf{restoration decoder} aims to reverse the degradation induced by the windshield and mirrors the encoder with five transposed convolution blocks, each with a transposed convolution layer, three batch normalization layers, and a merging node that incorporates high-fidelity information from the encoder. The number of filters is halved in each successive decoder block and the concatenated tensor is smoothed by two convolution layers with stride equal to one.

The \textbf{segmentation decoder} targets on the pixel-wise classification and also consists of five transposed convolution blocks. The high-fidelity information from the encoder is merged with the corresponding restoration layer before entering the concatenation layer in the decoder. This decoder coupling is supposed to enhance the segmentation performance against optical aberrations.

The downstream \textbf{residual encoder} for the regression of the Zernike coefficients comprises five ResNet~\cite{ResNet} cells. Each cell contains two convolution layers and two batch normalization layers, as well as a residual connection to merge the convolved signal with the input signal in order to alleviate the vanishing gradient problem~\cite{ResNet, ResNet_VGP, ResNet_VGP_norm}. Finally, a max pooling layer is employed for downsampling. Subsequent to the ResNet cells, the signal is flattened and passed through five dense layers with batch normalization and dropout for regularization.

\begin{figure*}[!t]
    \centering
    \captionsetup[subfigure]{oneside, margin={0.7cm, 0cm}}
    \begin{subfigure}[t]{0.24\linewidth}
        \centering
        \includegraphics[width=1.\linewidth]{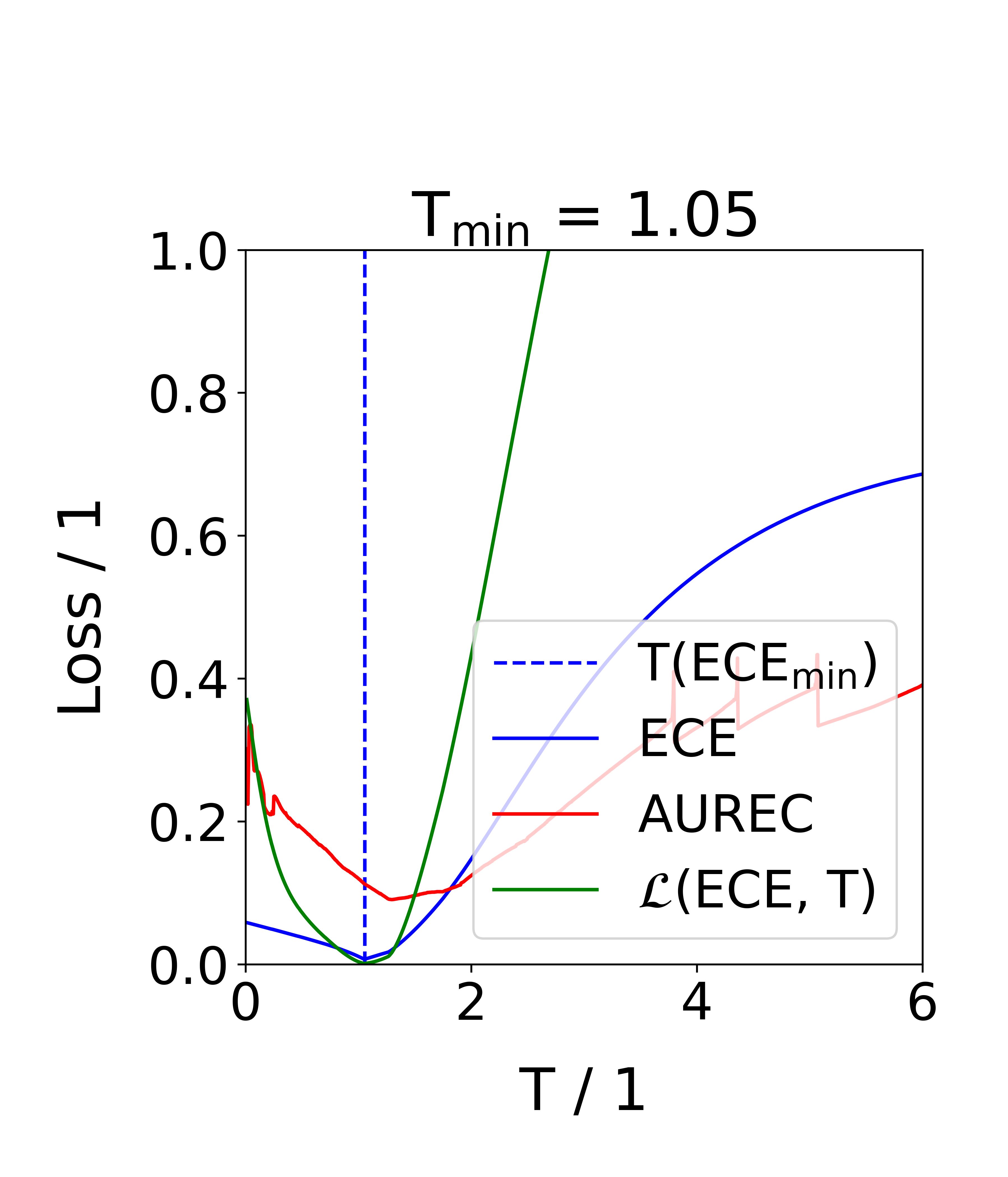}
        \caption{}
        \label{fig:loss_function_analysis_i1}
    \end{subfigure}
    \hfill
    \captionsetup[subfigure]{oneside, margin={0.7cm, 0cm}}
    \begin{subfigure}[t]{0.24\linewidth}
        \centering
        \includegraphics[width=1.\linewidth]{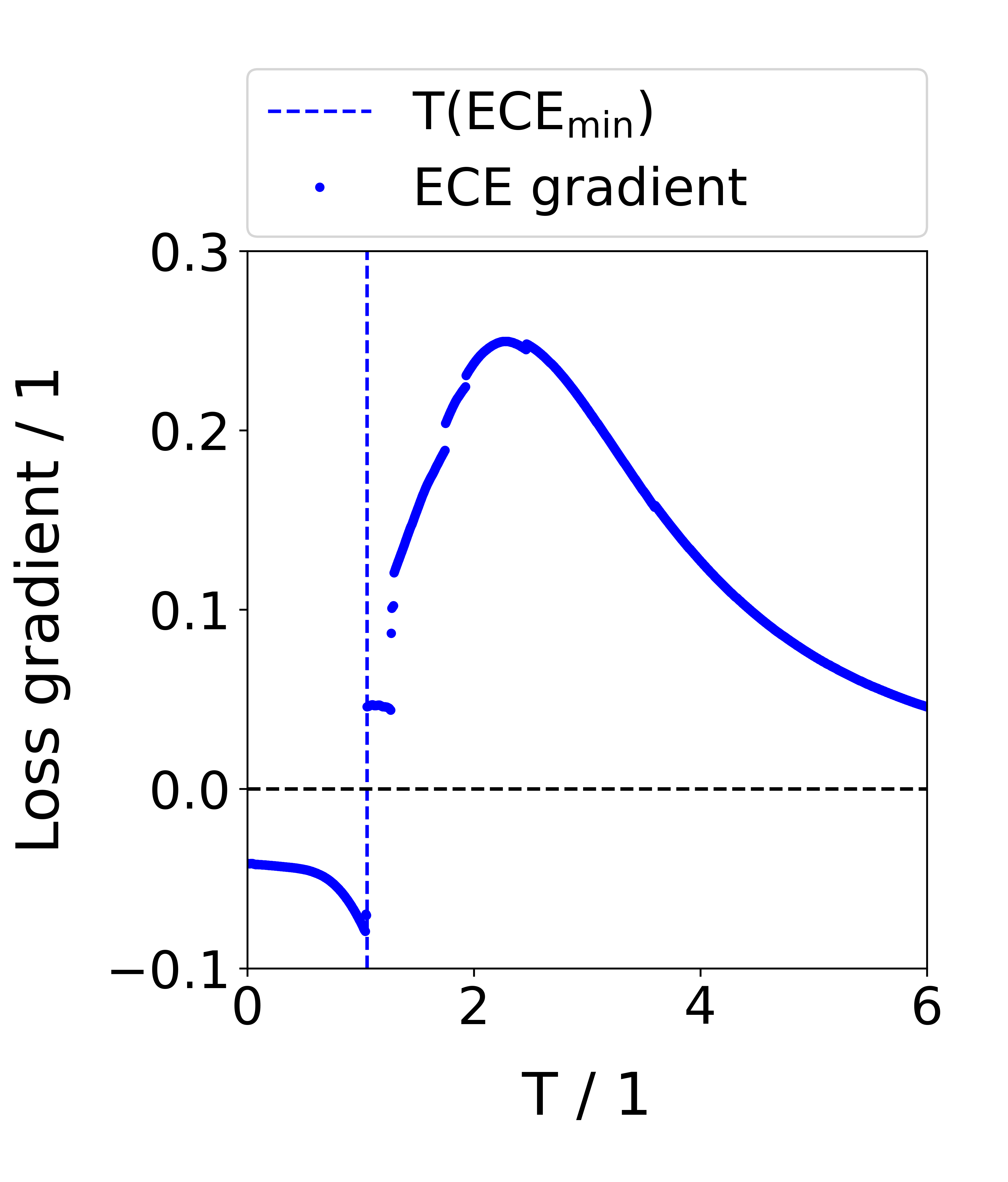}
        \caption{}
        \label{fig:loss_function_analysis_i2}
    \end{subfigure}
    \hfill
    \captionsetup[subfigure]{oneside, margin={0.7cm, 0cm}}
    \begin{subfigure}[t]{0.24\linewidth}
        \centering
        \includegraphics[width=1.\linewidth]{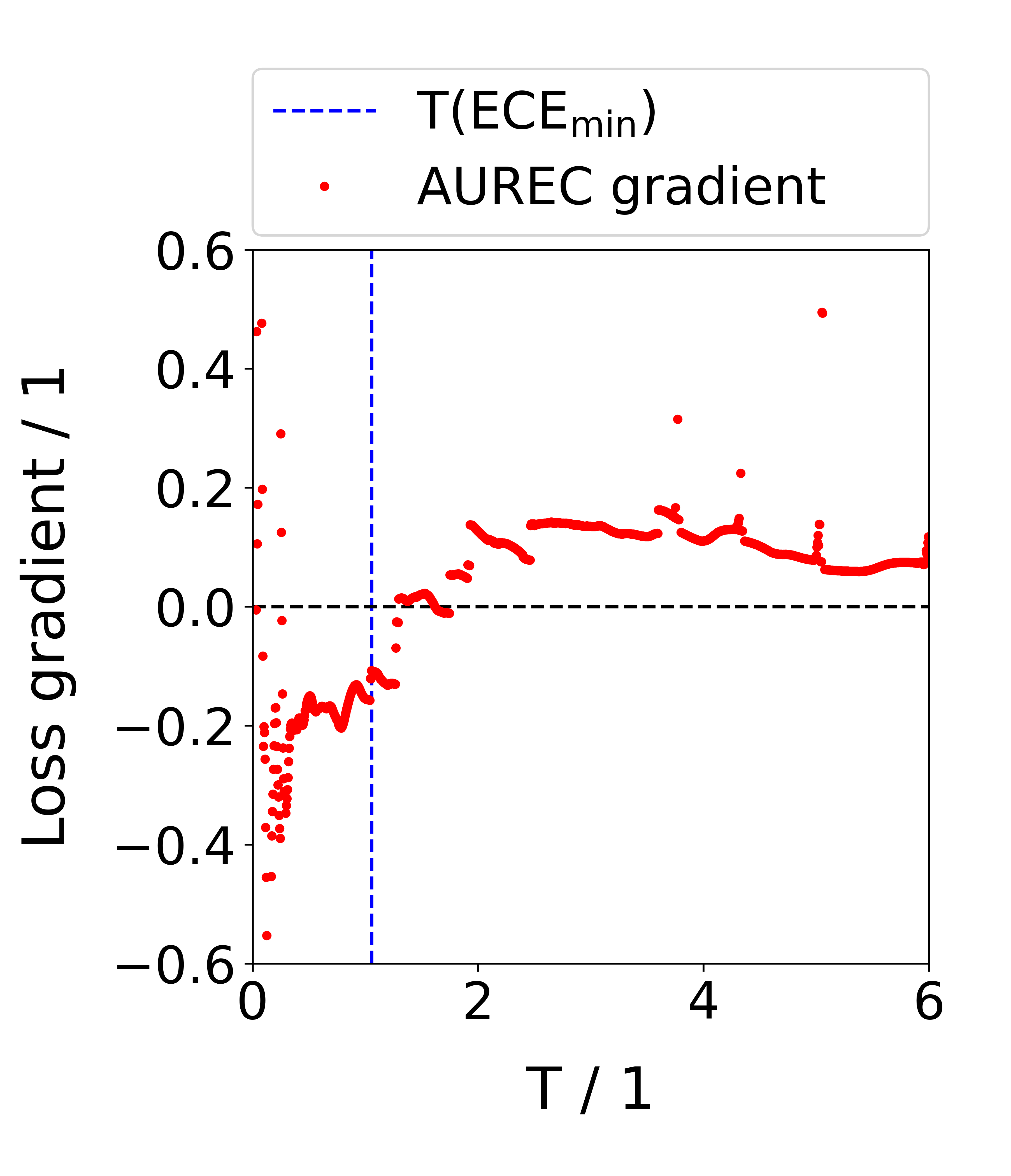}
        \caption{}
        \label{fig:loss_function_analysis_i3}
    \end{subfigure}
    \hfill
    \captionsetup[subfigure]{oneside, margin={0.85cm, 0cm}}
    \begin{subfigure}[t]{0.24\linewidth}
        \centering
        \includegraphics[width=1.\linewidth]{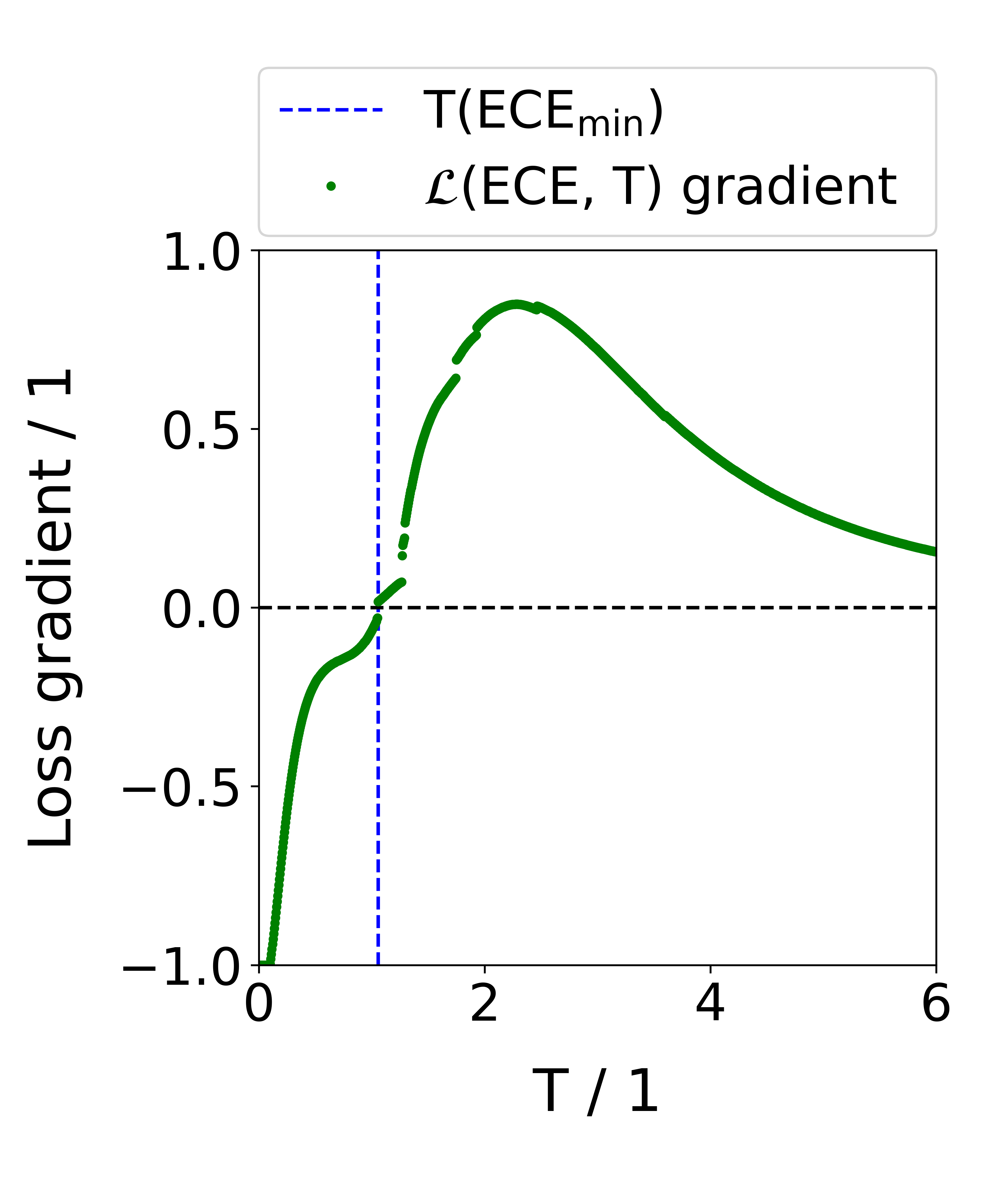}
        \caption{}
        \label{fig:loss_function_analysis_i4}
    \end{subfigure}
    \caption{Loss function study for the PIPTS calibration network. The loss is indicated for a random instance as a function of the calibration temperature in (a) with~$\beta_{s} = 1000$ and $N_{b} = 10$. The smoothed ECE measure is plotted as a blue line and the corresponding gradient is visualized in (b). The discontinuity at the optimal temperature~$\mathrm{T_{min}}$ indicates the need for an additional modulation function. The gradient of the total loss function, containing the modulation function and the temperature regularization term, is visualized in (d). It can be concluded that the total loss~$\mathcal{L}$ is sufficiently continuous differentiable~($C^{1}$) for backpropagation. Furthermore, the gradient of the AUREC is plotted in (c) as a function of the calibration temperature. The number of peaks indicates, that the smoothing of the AUREC loss function by the softmax function was insufficient to ensure continuity. Hence, the AUREC loss function is inadequate for backpropagation and for neural network training respectively.}
   \label{fig:loss_function_analysis}
\end{figure*}

\subsection{Dataset and preprocessing}
The network will be trained on the A2D2~\cite{A2D2} dataset because it provides pixel-wise labels for high resolution images with $1208 \times 1920$~pixels that were captured without a windshield, a necessary condition for generating synthetically degraded images in order to enrich the data heterogeneity. Only the images from the narrow-view front-center camera (Sekonix SF3325-100~\cite{A2D2}) are used due to its sensitivity to optical aberrations. The dataset comprises $5400$ training images and $1350$ test images, corresponding to a $20\%$ test split. To ensure comparability, the $38$-class ground truth annotations of the A2D2~\cite{A2D2} dataset are mapped to the $19$-class labeling taxonomy of the Cityscapes~\cite{Cordts2016Cityscapes} dataset. For the multi-task network, as for the PIPTS calibration network, the input and output tensors are normalized to zero mean and unit variance in order to equalize the dissimilarity of feature units.

\subsection{Training configuration}
The training was distributed across two Nvidia RTX A6000 GPUs with 48GB memory. A learning rate schedule reduced the learning rate by a factor of $10$ if the validation loss plateaued for the last $40$~epochs, and the training was terminated if no improvement occurred over the last $100$~epochs.

\subsection{Loss Function}
The multi-task network is trained by utilizing a customized loss function with three components:
\begin{enumerate}
    \item{\textbf{Semantic segmentation loss:} Negative log-likelihood (aka. cross-entropy) loss with a focal modulation term $(1 - \hat{p}_{i})^\gamma$, which enforces a focus of the learning process on classes that are hard to learn~\cite{Focal_loss}. Additionally, class balancing is applied to equalize the representation of different classes within the dataset~\cite{dreissig2023calibration}. The weighting factors~$\tau_{i}$ for each class~$i$ are calculated based on their frequencies in the dataset:
    \begin{equation}
        \tau_{i} = \dfrac{\log{\left(1.1 + \dfrac{c_{i}}{N}\right)^{-1}}}{\sum \limits_{i=1}^{N_{c}} \log{\left(1.1 + \dfrac{c_{i}}{N}\right)^{-1}}}\;,
    \end{equation}
    where $c_{i}$ denotes the number of instances for class~$i$, $N$ characterizes the total number of pixels within the A2D2 dataset, and $N_{c}$ specifies the number of classes.}
    \item{\textbf{Restoration loss:} L1-norm between the unperturbed images and the restored images~\cite{L1_over_L2}.}    
    \item{\textbf{Zernike regression loss:} L2-norm between the predicted and the ground truth Zernike coefficient vectors.}
\end{enumerate}
The restoration loss and the Zernike regression loss add to the total loss by considering an associated weighting factor, determined via hyperparameter tuning.

\subsection{Hyperparameters}
In addition to the loss term weighting factors, there are four other hyperparameters to tune:
\begin{enumerate}
    \item The weighting factor of the Kernel-Orthonormality-Regularizer~(KOR) term (see Section~\ref{sec:Model_regularization}), which is added linearly to the loss function as a penalty.
    \item The learning rate, which determines the increment in the optimization process.
    \item The focal loss exponent~$\gamma$.
    \item The batchsize, which relates to the number of images processed before the trainable variables are updated. When employing a large batchsize, the model's quality often degrades, particularly in terms of its generalization capabilities. Models with large batchsizes are prone to reaching sharp minima in the loss landscape, which are generally associated with reduced generalization performance. Conversely, small batchsizes tend to converge to flatter minima due to the stochasticity induced into the gradient estimates~\cite{Large_batch_sizes}.
\end{enumerate}

\subsection{Activation functions}
The Gaussian Error Linear Unit~(GELU)~\cite{GELU} is used as an activation function for the shared encoder, the segmentation decoder, and the restoration decoder. GELU is defined as $x \phi(x)$, where $\phi(x)$ is the standard Gaussian cumulative distribution function. The GELU activation function returns a likelihood-based output, unlike the Rectified Linear Unit~(ReLU)~\cite{RELU}, which simply gates the input according to the sign. Generalizing from monotonic~(e.g.~ReLU) to non-monotonic~(e.g.~GELU) activation functions can increase a neuron's discriminative capacity as has been demonstrated for the XOR problem~\cite{Non_linear_activation_function}. For the regression task, the Exponential Linear Unit~(ELU)~\cite{ELU} is employed in the ResNet cells and the dense layers. Moreover, the activation functions used in the output layers are specifically tailored for their respective tasks. The final layer of the segmentation head applies a softmax activation function. The restoration decoder implements the ReLU as an activation function for the output layer. Furthermore, the regression output is obtained by a linear activation function.

\subsection{Model regularization}
\label{sec:Model_regularization}
In order to enhance the generalization capabilities of the model, Kernel-Orthonormality-Regularization~(KOR)~\cite{KOR} is utilized. KOR penalizes orthonormality violations of the convolutional kernel matrices leading to reduced feature redundancy, which enriches the information capacity of the latent space embedding and boosts the model generalizability. To implement this, the convolutional kernel tensor is reshaped to a 2D-kernel matrix maintaining the innermost dimension~(number of output channels). Afterwards, the Gramian matrix is computed from the kernel matrix and the Frobenius norm is used for quantifying the residuals w.r.t.\ the identity matrix. The Frobenius norm corresponds to the Euclidean norm of the vector of eigenvalues of the matrix.

\subsection{PIPTS calibration network}
The PIPTS approach is implemented by a secondary, downstream CNN model, which utilizes the predicted logit tensor~$\boldsymbol{\omega}$ from the semantic segmentation head of the baseline model and the estimated Zernike coefficient vector~$\vec{\alpha}$ from the restoration head of the baseline model to predict an instance-wise temperature~$\mathrm{T_{cal}}$ for online calibration.

Using the flattened logits tensor~$\boldsymbol{\omega}$ and the Zernike coefficient vector~$\vec{\alpha}$ to directly determine the temperature did not achieve satisfying results. By encoding the logit tensor~$\boldsymbol{\omega}$ with another CNN before concatenating it with the Zernike coefficient vector~$\vec{\alpha}$ we achieved superior results. From this we hypothesize that the spatial distribution of logits in the image plays an important role for the calibration quality, as the spatial distribution of objects of different semantic classes (e.g., sky, persons etc.) also exhibits spatial features (e.g. the sky is up). By varying the number of encoder blocks we found that the number seven to be the best compromise between calibration quality and efficiency.

For the PIPTS model we will utilize the ECE directly as a loss function. The ECE metric is not differentiable because it implicitly relies on a counting operation for the confidence binning. Consequently, the ECE can not be used as a loss function a priori. We will utilize a mathematical trick to smooth the ECE metric such that it becomes continuous~(differentiability class $C^{0}$). The trick consists of employing the continuous softmax function with a large exponential scaling factor~($\beta_{s} = 1000$) in places where discontinuous operations are used, e.g., replacing the argmax operation. This will result in a differentiable function but it is not guaranteed that the derivative is continuous as well. In order to establish continuous differentiability~($C^{1}$), the smoothed ECE function is modulated.

Loss modulation is applied in order to raise the discontinuity in the ECE gradient around zero. For that reason, the gradient is modulated by the sigmoidal function~${f'(x ;\eta) := \tanh^2{(\eta x)}}$. As a result, the loss function is modulated by the function~$f(x ;\eta) := x - \eta^{-1} \tanh{(\eta x)}$, which imposes an inflection point around $x=0$. The hyperparameter~$\eta$ has been determined experimentally by hypertuning and amounts to~$\eta = 50$.

\begin{figure*}[!t]
   \centering
   \includegraphics[width=1.\linewidth]{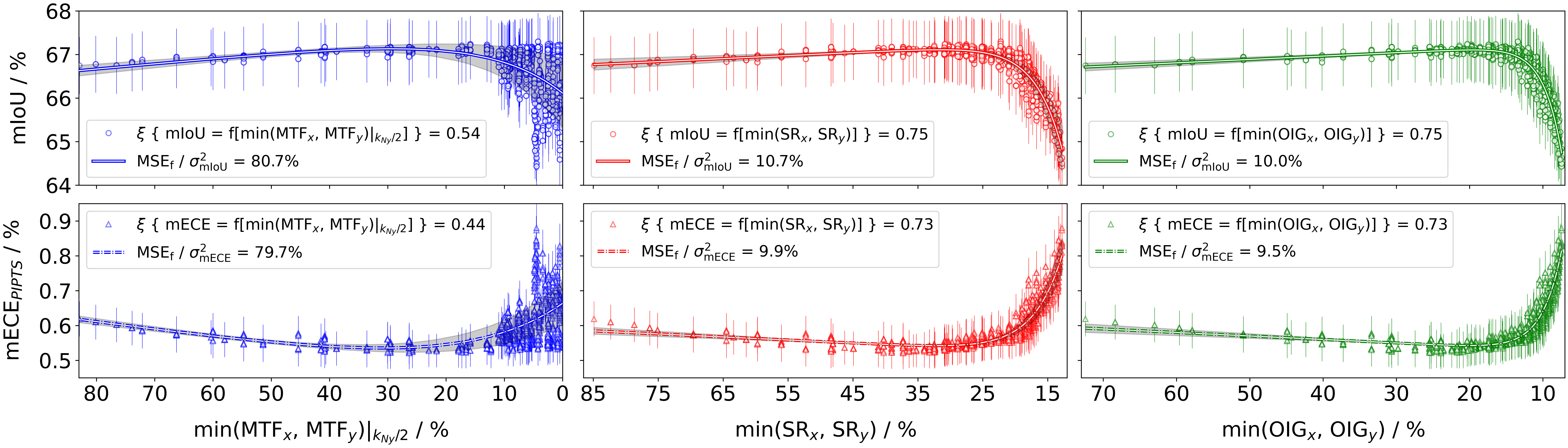}
   \caption{The dependency of the mIoU (upper row) and the mECE (lower row) on the MTF at half Nyquist frequency (left column), the Strehl ratio (middle column) and the OIG (right column) is plotted. The Strehl ratio and the OIG demonstrate a superior correlation to the mIoU and the mECE in terms of the Chatterjee rank correlation measure than the MTF at half Nyquist frequency. As a consequence, the regression function from Equation~(\ref{eq:model_function}) also fails to capture the non-existing relationship in the large-aberration regime but it performs well for the Strehl ratio and the OIG, which is quantitatively measured by the ratio of the Mean Squared Error~(MSE) over the variance~($\sigma^2$), referred to as the unexplained variance component.}
   \label{fig:Chatterjee_summary_plot}
\end{figure*}

Finally, Temperature regularization is applied in order to penalize predictions close to the temperature scaling pole at~$T=0$. The regularization term for the loss gradient is chosen as~$g'(x ;\kappa) := \tanh^2{(\kappa x)} - 1$. Considering the constraint that the regularization term for the loss function has to be positive-definite, the regularization term for the loss function is given by~$g(x ;\kappa) := - \kappa^{-1} \left( \tanh{(\kappa x)} - 1 \right)$. The hyperparameter~$\kappa$ has been quantified empirically by hypertuning to~$\kappa = 8$.

The hypertuning of the hyperparameters~$\eta$ and~$\kappa$ is supposed to remain valid for different datasets and neural network architectures, as long as the normalization of the input and output is maintained and the characteristics of the ECE metric are not modified. In particular, the number of bins will heavily affect the value of the hyperparameters because the number of bins determines the number and magnitude of the discontinuities in the ECE metric.

As a result, the loss function for the PIPTS training is given by:
\begin{align}
\begin{split}
    \mathcal{L}(\mathrm{ECE},\;\mathrm{T}) &:= f(\mathrm{ECE}) + g(\mathrm{T}) \\[2pt]
    \Rightarrow \mathcal{L}(\mathrm{ECE},\;\mathrm{T}) &\;= \mathrm{ECE} - 0.02 \tanh{(50 \;\mathrm{ECE})} \\[2pt]
    &\;- 0.125 \left( \tanh{(8\;\mathrm{T})} - 1 \right)\;,
\end{split}
\label{eq:loss_modulation}
\end{align}
and the gradient w.r.t.\ the weights~$\theta_{k}$ is modulated by:
\begin{align}
\begin{split}
    \dfrac{\partial \mathcal{L}}{\partial \theta_{k}} &\overset{(\ref{eq:loss_modulation})}{=} \left( \dfrac{\partial f}{\partial \mathrm{ECE}} \cdot \dfrac{\partial \mathrm{ECE}}{\partial \mathrm{T}} + \dfrac{\partial g}{\partial \mathrm{T}} \right) \dfrac{\partial \mathrm{T}}{\partial \theta_{k}} \\[2pt]
    \Rightarrow \dfrac{\partial \mathcal{L}}{\partial \theta_{k}} &= \tanh^2{(50\;\mathrm{ECE})} \cdot \dfrac{\partial \mathrm{ECE}}{\partial \mathrm{T}} \dfrac{\partial \mathrm{T}}{\partial \theta_{k}} \\[2pt]
    &+ \left( \tanh^2{(8\;\mathrm{T})} - 1 \right) \dfrac{\partial \mathrm{T}}{\partial \theta_{k}}\;.
\end{split}
\label{eq:loss_modulation_derivative}
\end{align}
Figure~\ref{fig:loss_function_analysis} visualizes the development process of constructing the loss modulation function~$f$ and the temperature regularizer~$g$ presented in Equation~(\ref{eq:loss_modulation}).

\section{Experiments and results}
The results of our contribution are split into three parts. First, we will elaborate on the performance of the baseline multi-task network for semantic segmentation. Secondly, the non-linear correlation between the mIoU and mECE versus the optical quality in terms of different metrics is studied by utilizing the Chatterjee correlation measure. Finally, the performance gain of the PIPTS calibrator is analyzed in comparison to the state-of-the-art PTS approach.

\subsection{Multi-task network performance and the insufficiency of the half-Nyquist criterion}
\begin{figure}[!b]
    \centering
    \begin{minipage}{.49\linewidth}
        \centering
        \includegraphics[width=\linewidth]{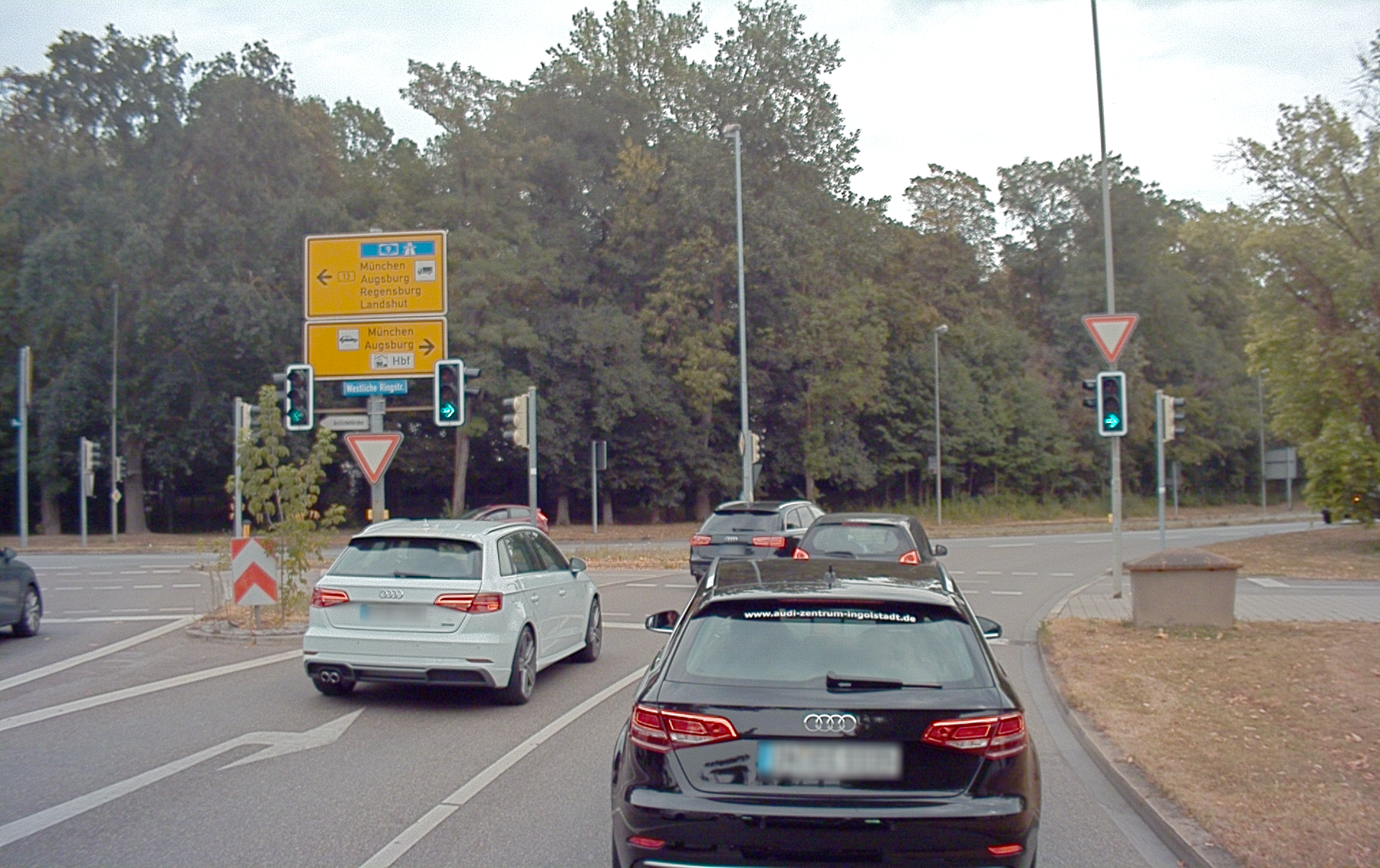}
    \end{minipage}
    \hfill
    \begin{minipage}{0.49\linewidth}
        \centering
        \includegraphics[width=\linewidth]{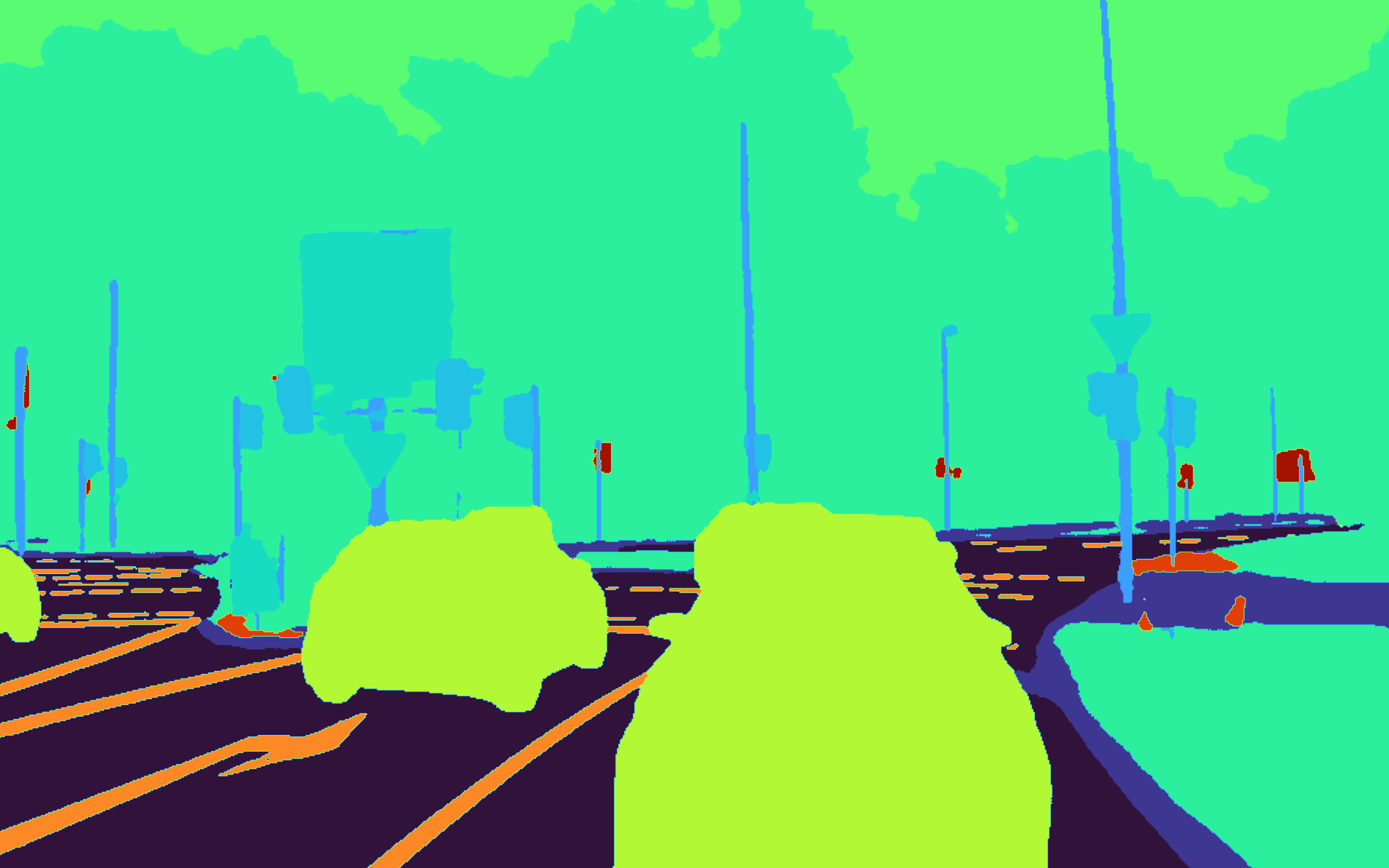}
    \end{minipage}
\caption{Segmentation map predicted by the multi-task network for a random instance of the A2D2~\cite{A2D2} dataset.}
\label{fig:baseline_performance}
\end{figure}

The performance w.r.t.\ the target application, semantic segmentation of the A2D2 dataset, is quantitatively evaluated in terms of the mIoU. Across the entire test dataset, the peak average mIoU evaluates to $\mathrm{mIoU_{test}} = (67.3 \pm 0.7)\%$. Furthermore, the segmentation performance is qualitatively visualized by a random instance in Figure~\ref{fig:baseline_performance}.

Moreover, Figure~\ref{fig:Chatterjee_summary_plot} depicts the mIoU and the mECE as a function of different optical metrics. It is evident that the peak performance is not given in the absence of optical aberrations~(diffraction-limited case), which might be counterintuitive at first glance. The mIoU is maximized for instances that reflect the mean-level of optical aberrations within the training dataset. As a consequence, it is of paramount importance to incorporate the optical aberrations within the perception chain proportionally to their occurrence in part-level measurements. By doing so, the augmented training dataset will be centered at the expected optical quality of the produced perception chain, and the mIoU as well as the mECE will be implicitly tuned for this aberration scenario.

This conclusion aligns well with previous work in the field of deep optics~\cite{Deep_optics_I, Deep_optics_II, Deep_optics_III}, where the perception chain is holistically optimized alongside the neural network training. This is done by constructing a differentiable, physics-based optics model and by assigning the corresponding optical parameters as trainable hyperparameters. The contributions in this field~\cite{Deep_optics_I, Deep_optics_II, Deep_optics_III} strongly indicate that optical quality is not all what you need. To the best of our knowledge, this result has only been shown with respect to the target application performance, e.g., image classification, depth estimation, 3D object detection etc. With our work, we demonstrate that this effect also holds true in terms of the neural network calibration performance.

In order to quantitatively assess this effect we postulate a regression function~$f$:
\begin{equation}
    f\left(x\;;\;\vec{\beta}\right) := \beta_{1} \exp{\left( \beta_{2} \left( x - \beta_{3} \right)\right)} + \beta_{4} x + \beta_{5}\;,
\label{eq:model_function}
\end{equation}
which is supposed to capture the mIoU and mECE performance as a function of the optical quality. The first term accounts for the exponential decay of the mIoU in the large-aberration regime and the exponential increase of the mECE, respectively. Furthermore, the last term denotes an ordinate offset. Finally, the performance gain from the low-aberration regime to the mean-aberration regime is captured by a linear term and the corresponding slope measures the magnitude of the aforementioned effect. In addition, the global extremum of the regression function quantifies the mean optical quality of the training dataset.

The regression function~$f$ is parameterized by the coefficient vector~$\vec{\beta}$ and the combined uncertainty~$\sigma_{c}$ will be determined by applying the multivariate law of uncertainty propagation~\cite{GUM_compliance}:
\begin{equation}
    \resizebox{1.\linewidth}{!}{$
    \dfrac{\sigma_{c}(x_{i})}{k_{\mathrm{\nu_{eff}}}} = 
    \sqrt{
        \left(\vec{\nabla}_{\vec{\beta}}f\right)^{T}\Bigr|_{x_{i}}
        \left[ \begin{array}{ccc}
           \sigma^2_{\beta_{1}} & \cdots & \rho_{\beta_{1},\beta_{d}} \sigma_{\beta_{1}} \sigma_{\beta_{d}} \\
           \vdots & \ddots & \vdots \\
           \rho_{\beta_{1},\beta_{d}} \sigma_{\beta_{1}} \sigma_{\beta_{d}} & \cdots & \sigma^2_{\beta_{d}}
        \end{array} \right]
        \vec{\nabla}_{\vec{\beta}}f\Bigr|_{x_{i}}
    }\;.$}
\label{eq:model_function_unc}
\end{equation}
The covariance matrix for the parameters~$\beta_{i}$ is calculated by a Monte-Carlo study~\cite{Monte_Carlo} considering the batch-wise standard deviation of the mIoU and the mECE respectively. In total $N = 1000$ regression curves were calculated, wherefore the extension factor~$k_{\nu_{\mathrm{eff}}}$ is given by $k_{995} = 1.96$ for a confidence level of $95\%$~\cite{GUM, GUM_II}. Finally, the symmetrical interval spanned by the combined uncertainty~$\sigma_{c}$ determines the confidence bands around the regression function and is illustrated in Figure~\ref{fig:Chatterjee_summary_plot} in gray.

The explanatory power of the postulated regression function~(\ref{eq:model_function}) is further quantified in terms of the unexplainable variance, which is given by the ratio of the mean squared error (MSE)~\cite{Oxford_Dictionary_of_Statistics} over the variance of the dataset itself. Statistically, the MSE measures the ensemble spread around the regression line. Benchmarking regression models according to the unexplainable variance is favorable because it effectively measures how well the regression model outperforms the naive estimate given by the arithmetic mean~\cite{Wolf_Metrologia}. In addition, if the complexity of the regression model is increased the degrees of freedom in the MSE computation decreases, which acts as a penalty for more complex regression models like in Ridge regression~\cite{Ridge_regression_I, Ridge_regression_II, Ridge_regression_III}. The quantitative values for the unexplainable variance are presented in Figure~\ref{fig:Chatterjee_summary_plot} and indicate that the model function does not significantly outperform the arithmetic mean in case of the MTF. On the other hand, the model function is superior for the Strehl ratio and the OIG. Consequently, the coefficient vector~$\beta$ can be interpreted as a sensitivity vector in case of the Strehl ratio and the OIG. Of special importance are~$\beta_{4}$ for the linear performance drift in the low-aberration regime and the coefficients~$\beta_{1}$ and~$\beta_{2}$ for the exponential decay in the large-aberration regime. These sensitivity coefficients can be employed for determining valid operational domains for the perception chain and for comparing the robustness across different neural network architectures.

The main reason why the proposed regression function does not sufficiently capture the variability of the mIoU and the mECE as a function of the MTF lies in the lack of correlation. For all three optical measures of interest -- the MTF at half Nyquist frequency, the Strehl ratio as well as the OIG -- the Chatterjee rank correlation measure~$\xi$ is computed. The correlation is most evident for the Strehl ratio and the OIG~{($\xi_{1331} = 0.75$)}, whereas the MTF at half Nyquist frequency is a significantly worse indicator for the AI performance~{($\xi_{1331} = 0.54$)}.

\begin{figure*}[t!]
    \centering
    \includegraphics[width=\linewidth]{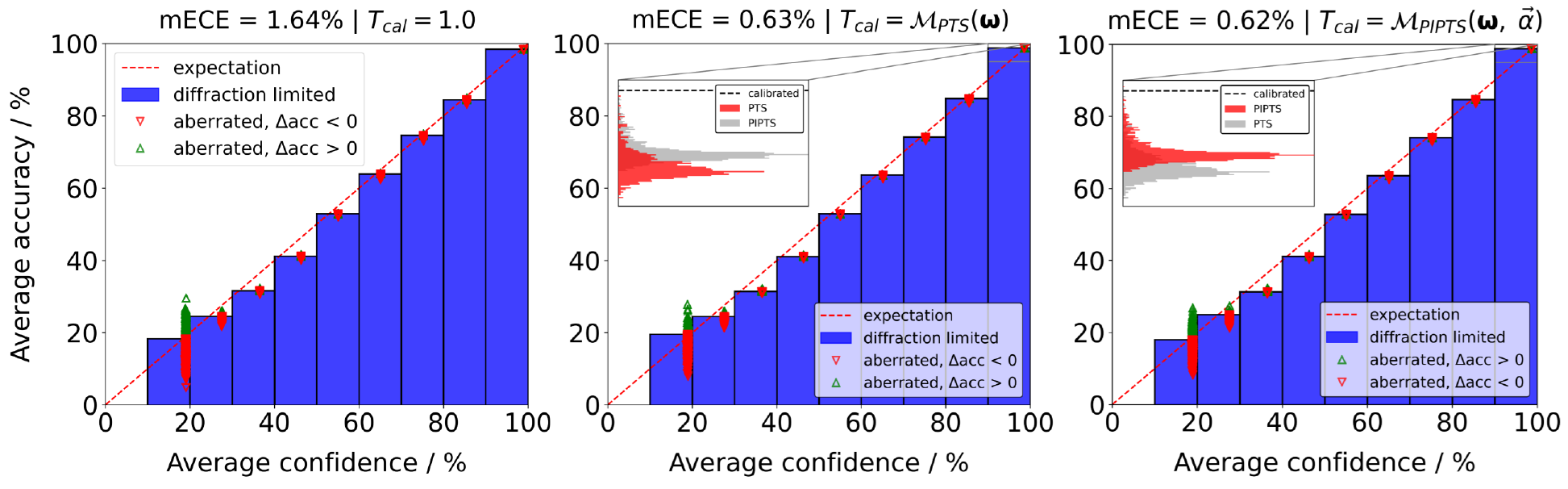}
    \caption{The reliability diagrams for the multi-task network are presented if calibrated by: (left) TS, (middle) PTS, and (right) PIPTS. If TS is directly contrasted against PTS, then the performance gain in terms of the mECE amounts to over $1\%$, which is tremendous. The calibration performance can be further boosted if a physical inductive bias is included into the PTS architecture. This benefit becomes significant if dataset shifts due to optical aberrations in the perception pipeline are induced. The impact of different perturbation scenarios is visualized by red~(average bin accuracy is decreased) and green~(average bin accuracy is increased) triangles. For the bin with the highest cardinality w.r.t.\ the subset of predictions, the distribution of the calibration error in terms of the difference between the average confidence and average accuracy is highlighted by an auxiliary plot. If PTS is contrasted against PIPTS, it is noticeable that the calibration bias is significantly reduced by adding additional information about the magnitude of the dataset shift. Furthermore, a slight reduction of the variance of the calibration error distribution can be observed.}
\label{fig:reliability_diagrams}
\end{figure*}

Nevertheless, it is evident, that the MTF at half Nyquist frequency fits the regression function well in the low-aberration regime but the huge spread in the codomain for the large-aberration regime leads to a wide confidence band around the global trend. The correlation of the MTF breaks down for severe optical aberrations because of the monotonicity violation in the large-aberration regime. As long as the MTF function is monotonically decreasing, the value of the MTF function at half Nyquist frequency will correlate with the area enclosed by the MTF curve. Since the area under the MTF curve, which equals the Strehl ratio if normalized by the diffraction-limited case, shows a robust correlation to the mIoU across the entire spatial frequency domain, the MTF at half Nyquist frequency is a valid optical performance indicator as long as the MTF curve is monotonically decreasing. As the optical aberrations in the automotive industry are typically too severe in magnitude to satisfy the monotonicity constraint, the MTF at half Nyquist frequency is not a suitable measure for safeguarding AI-based autonomous driving algorithms against optical perturbations. Consequently, system MTF requirements evaluated at half-Nyquist frequency should be considered as invalid for the large-aberration regime.

\subsection{PIPTS calibration quality}
The calibration performance is visualized by the reliability diagram in Figure~\ref{fig:reliability_diagrams} for the baseline model calibrated with TS. As the multi-task network is inherently well calibrated for the diffraction-limited case by utilizing class-balancing and loss-focusing during training, the optimal temperature according to TS is given by $T_{cal} = 1.0$, which corresponds to an identity mapping of the confidences. In addition, Figure~\ref{fig:reliability_diagrams} visualizes the reliability diagrams for the multi-task network after applying PTS and PIPTS, respectively. It is evident that the instance-wise calibrators significantly outperform TS by over~$1\%$. This highlights the gain in expressive power provided by the superior information capacity of the post-hoc CNN-based calibrators~\cite{PTS}. The physical inductive bias of PIPTS results in an additional mECE boost of roughly $100\;\mathrm{ppm}$ for the diffraction-limited case. If aberrations are considered, then the knowledge about the dataset shift magnitude successfully counteracts the calibration bias. In addition to accuracy improvements, PIPTS also slightly improves the calibration precision, which is notable by narrower bin-wise calibration error distributions.

\begin{figure*}[!t]
    \centering
    \subfloat[The net performance boost of the PIPTS calibrator is indicated as a function of the Strehl ratio.\\]{%
    \includegraphics[clip, width=0.8\linewidth]{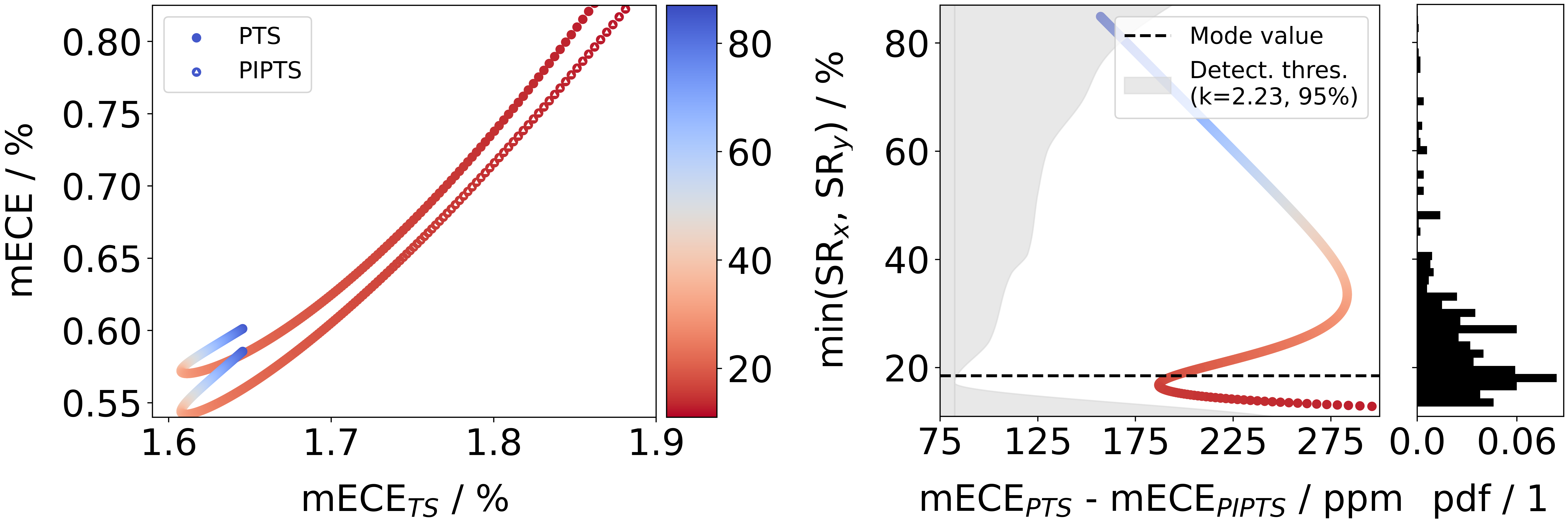}
    \label{fig:TS_vs_PIPTS_mECE_vs_SR}}
    \\[\baselineskip]
    \subfloat[The net performance boost of the PIPTS calibrator is indicated as a function of the OIG.]{%
    \includegraphics[clip, width=0.8\linewidth]{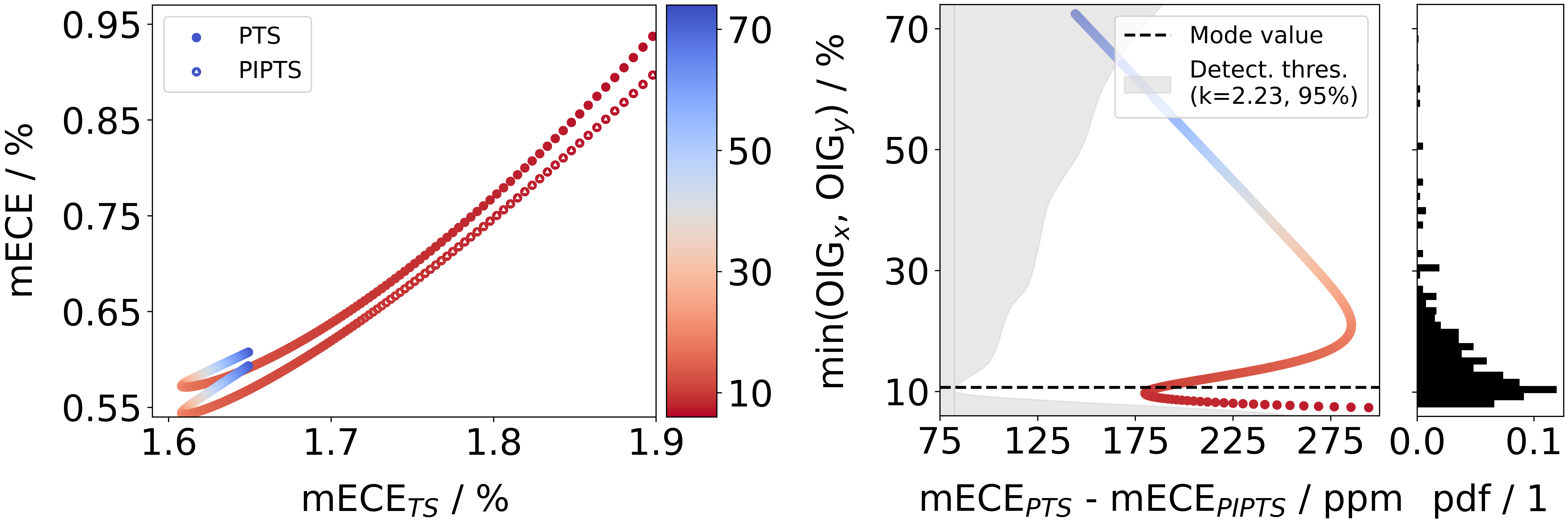}
    \label{fig:TS_vs_PIPTS_mECE_vs_OIG}}
    \vspace{0.2cm}
    \caption{On the left-hand side, the mECE curves for the PTS and PIPTS calibrator are plotted versus the TS calibration performance. The colorbar indicates the aberration magnitude in terms of the (a)~Strehl ratio and (b)~OIG. The curves are obtained by averaging over 11 post-hoc models in a Deep Ensemble fashion. The graph in the middle shows the performance boost of PIPTS in comparison to PTS over the aberration magnitude as well as the corresponding detection threshold on a $95\%$ confidence level in gray. On the right-hand side, the histogram of the augmented dataset distribution in terms of the (a)~Strehl ratio and (b)~OIG is visualized. It is evident that the mode value of the (a)~Strehl ratio distribution and (b)~OIG distribution correlates with the local minimum in the performance boost curve. In summary, the performance boost induced by the physics prior in PIPTS is significant for the mean- and large-aberration regime if PSF-based optical measures -- in particular the Strehl ratio or the OIG -- are considered.}
\end{figure*}

A valid question might be raised about how significant this performance boost is. In order to tackle this question, we utilized the Deep Ensemble approach~\cite{Deep_ensembles}. An ensemble of 11~PIPTS models has been trained with the same hyperparameters~(congruent loss function landscape in the parameter space) but random and hence different weights initialization. Each ensemble member is used to calibrate the baseline model and the robustness analysis presented in Figure~\ref{fig:Chatterjee_summary_plot} is repeated for the well-correlating optical metrics, the Strehl ratio and the OIG. Consequently, the violation of the calibration condition~\cite{Wolf_GCPR24} under optical aberrations can be quantified by extracting the regression function parameters~$\beta_{q}$. Subsequently, the mean performance boost across all ensemble members and the corresponding standard deviation for the mean are evaluated and visualized in Figure~\ref{fig:TS_vs_PIPTS_mECE_vs_SR} and \ref{fig:TS_vs_PIPTS_mECE_vs_OIG}. Utilizing the standard deviation of the mean calibration performance in terms of the mECE as a predictive uncertainty estimator for the PIPTS calibrator implicitly neglects the aleatoric uncertainty component. If it is intended to include this component into the uncertainty estimation regarding the mECE, second order distributions have to be considered. This could be done by extending the output space of the PIPTS calibration network such that the conditional uncertainty of the temperature is predicted alongside the instance-wise calibration temperature. The conditional uncertainty can then be propagated to retrieve an uncertainty estimate for the mECE. Unfortunately, this approach would require us to adjust the loss function. If a Gaussian distribution is assumed to characterize the noise in the training dataset of the PIPTS calibrator, then the Gaussian negative log-likelihood loss~\cite{Gaussian_NLL} could be employed for training. The advantage of learning the conditional uncertainty alongside the instance-wise calibration temperature also comes with a downside, as the Gaussian negative log-likelihood loss relies on ground truth temperatures. As this information is not available a priori, the optimal temperature would need to be retrieved before training for every instance by minimizing the mECE. In order to avoid a supervised regression setup for the PIPTS calibration network training, we employ the modulated mECE directly as a loss function, which eliminates the need of ground truth temperatures. As a consequence, we will restrict our study to first order distributions and the PIPTS calibrator exclusively predicts the temperature. Neglecting the aleatoric uncertainty component for the evaluation of the PIPTS calibration performance is reasonable, as the aleatoric uncertainty component essentially measures the quality of the baseline segmentation model because of the noise that the baseline model generates in its predictions due to misclassifications. In order to avoid this dependency on the performance of the segmentation model, only the epistemic uncertainty estimate is used for assessing the predictive uncertainty of the PIPTS calibration network. The significance level of the PIPTS performance gain has been set to $95\%$ utilizing the Student's t-distribution and a corresponding extension factor of~$k_{10} = 2.23$ for an ensemble with~$\nu_{\mathrm{eff}} = 10$ degrees of freedom~\cite{GUM, GUM_II}. Here, the Student's t-distribution has to be considered, as the standard deviation is estimated based on a very limited number of degrees of freedom~\cite{GUM, GUM_II}.

Figure~\ref{fig:TS_vs_PIPTS_mECE_vs_SR} depicts the mECE curves for the PTS and PIPTS calibrator against the TS calibration results for the Strehl ratio. Those curves do not represent functional relationships because the calibration performance for all three post-hoc techniques is maximized for the mean-aberration regime. The PIPTS performance boost is not significant for the diffraction-limited case if a confidence level of $95\%$ is considered in the view of the Student's t-distribution. The performance boost increases almost linearly with the aberration magnitude until the mean-aberration regime is approached. As the optical degradations approach the level of the mean optical quality of the training dataset, given by the dataset centroid, the benefit of employing PIPTS decreases. Hence, the supplementary information about the Zernike coefficients is not as beneficial as for long-tail samples with fewer aberrations in terms of the Strehl ratio. This might be the case if a neural network trained on augmented data is used to infer information from synthetic data, which typically does not comprise optical aberrations of the windshield. On the other hand, in the large-aberration regime, the performance boost of PIPTS increases again after passing the mode value of the augmented dataset distribution. This scenario might occur if aging effects of the windshield are considered.

A similar outcome is observed if the OIG is considered as the optical target metric, as illustrated in Figure~\ref{fig:TS_vs_PIPTS_mECE_vs_OIG}. As a result, the PIPTS performance boost is significant on a $95\%$ confidence level for the mean- and large-aberration regime but not for the low-aberration regime close to the diffraction limit, as the physical prior does not add significant information within this domain.

The significant calibration performance boost of PIPTS in the order of $250\mathrm{ppm}$ needs to be considered in the light of the number of instances that occur in the lifetime of an autonomous driving car fleet. As a thought experiment, if a fleet of $10$ million cars is taken into account, which corresponds to the annual production volume of large car manufacturers, and a lifetime of $200\;\mathrm{Mm}$ is considered per car, then a calibration error reduction of $250\mathrm{ppm}$ corresponds to an increase in the safety margin of $500\;\mathrm{Gm}$. This perspective underpins the benefits of PIPTS for autonomous driving.

\begin{figure}[!t]
   \centering
   \includegraphics[width=1.\linewidth]{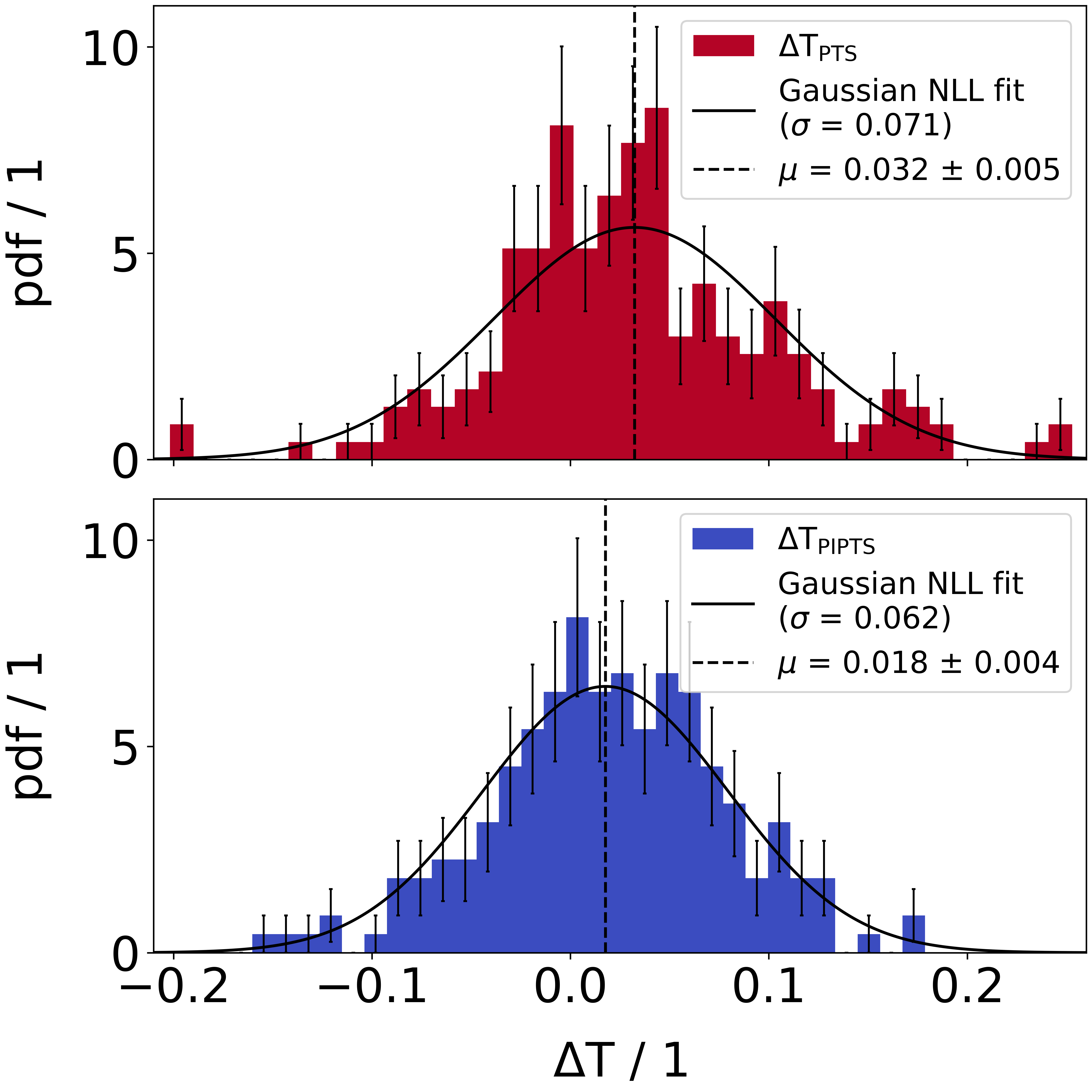}
   \caption{The temperature deviation~$\Delta T$ between the predicted temperature of the (top) PTS, (bottom) PIPTS calibrator and the optimal temperature is plotted as a histogram. Both distributions indicate a significant bias, which reflects the influence of the dataset shift on the post-hoc calibrators. The bias~$\mu$ is significantly reduced if the PIPTS calibrator is employed instead of the PTS model.}
   \label{fig:T_distribution}
\end{figure}

The performance boost of PIPTS -- generated by the physical inductive bias -- manifests itself as a slight distribution shift in the predicted temperature over the entire test dataset. As a toy-example, Figure~\ref{fig:T_distribution} depicts the histograms of the predicted temperature deviation for the aberration scenario:~$\alpha_{3} = -0.2\;\lambda$, $\alpha_{4} = 1.0\;\lambda$, and $\alpha_{5} = -1.0\;\lambda$. The temperature deviation is given as the difference between the instance-wise temperature predicted by the corresponding calibrator~(PTS or PIPTS) and the instance-wise optimal temperature, which is obtained by minimizing the mECE as a function of the temperature for each instance separately. The distributions show a non-zero bias, which is subject to the dataset shift induced by the optical aberrations. The bias is significantly reduced if the PIPTS calibrator is employed instead of the PTS calibrator. The parameters of the underlying Gaussian distribution~(bias~$\mu$ and standard deviation~$\sigma$) were determined by a negative log-likelihood fit and the corresponding parameter uncertainty is estimated by the inverse of the local curvature of the negative log-likelihood curve according to the Cramer-Rao-Frechet bound~\cite{Cramer,Rao,Frechet}.

\section{Benefits for autonomous driving}
PIPTS based on the logit tensor and the effective Zernike coefficients of the overall optical system is used to maintain the calibration quality of the predicted confidences of the multi-task network even under a wide variety of degradation-related dataset shifts due to internal~(e.g., ageing of the windshield, thermal effects (windshield heating, solar radiation, etc.)) and external factors~(e.g., weather influences, rock chips). This enhances the trustworthiness of the baseline multi-task network under optical aberrations. With this safety mechanism based on the optical quality, it is possible to dynamically monitor the hazard potential in real-time, thereby reducing situations that could jeopardize safety. As a result, the architectural enhancements introduced in this paper~(e.g., optical inductive bias for the PIPTS calibrator, coupled decoder head in the multi-task network, etc.) lead to superior robustness against aberration-related dataset shifts, which permits a wider definition of part-specific requirements and strengthens the real-world performance of the perception system.

The predicted, effective wavefront aberrations of the overall system can not only be used as an inductive bias for the PIPTS calibrator but also for end-of-line testing. The multi-task network enables the end-of-line testing by absorbing the non-linear interplay of the optical aberrations induced by the ADAS camera and the windshield into the information capacity of the multi-task model. The non-linear mapping of the part-level Zernike coefficient vectors of the ADAS camera and the windshield to the Zernike coefficient vector of the entire perception chain permits individual part testing at the supplier site, as it is essential for automotive industry processes~\cite{Braun_TM} according to the Vee-model~\cite{Vee_model}.

\section{Conclusion}
Our contribution manifests itself threefoldly. First, we quantitatively demonstrated that the Strehl ratio and the OIG outperform the MTF at half Nyquist frequency in terms of the Chatterjee rank correlation measure. Secondly, we showed experimental evidence on the superiority of PIPTS over PTS, which implies that incorporating physical priors to the PTS calibrator enhances its expressive power. Finally, we highlighted the benefits of the coupled decoder head for the Zernike coefficient prediction in the light of system-level requirements. The multi-task network provides a tool for capturing the non-linear mapping from the Zernike coefficient vectors of the ADAS camera and the windshield to the system-level wavefront aberration map. This enables the automotive industry to derive part-specific requirements as the verification strategy according to the Vee-model is demanding~\cite{Vee_model, Wolf_IMEKO2024}. In a nutshell, our contribution paves the way for establishing trustworthiness and robustness in AI-based autonomous driving functionalities by ensuring superior confidence calibration under optical aberrations in the perception chain and by providing a physical sound toolchain for deriving part-specific optical requirements.

\begin{IEEEbiography}[{\includegraphics[width=1in,height=1.25in,clip,keepaspectratio]{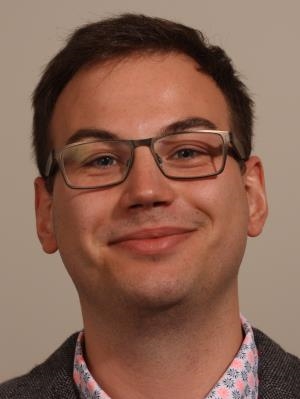}}]{Dominik Werner Wolf}
studied physics at the Karlsruhe Institute of Technology and at the Ruprecht-Karls-University of Heidelberg. His academic journey led him to CERN, where he conducted research on tune modulations at the LHC for his Bachelor's thesis and worked on novel beam stabilization techniques for his Master's thesis. At the OPMD lab of the University of Oxford, he contributed to the Mu3e detector development and the assembly strategy, showcasing interdisciplinary prowess. With over six years of industry experience in metrology and neural network robustness, he bridges theory and application seamlessly. His passion lies in computational physics, tackling problems across disciplines with relentless curiosity.
\end{IEEEbiography}

\vspace{-0.5cm}

\begin{IEEEbiography}[{\includegraphics[width=1in,height=1.25in,clip,keepaspectratio]{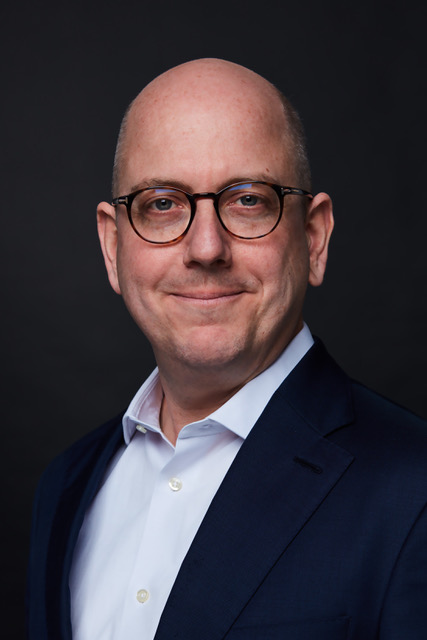}}]{Alexander Braun}
has 20 years of experience with optical technologies of all flavours, from fundamental quantum optics research to optical quality questions arising during assembly of 100.000 cameras per year. During his position as optical designer at Kostal (Automotive Tier 1) he was responsible for the optical quality of the camera ADAS, designing MTF test benches for all purposes. As a Professor for Physics at the University of Applied Sciences in Duesseldorf he is now looking at all the questions surrounding optical quality from a scientific point of view. Special research interests are the limits of current image quality algorithms and linking optical design to function performance.
\end{IEEEbiography}

\begin{IEEEbiography}[{\includegraphics[width=1in,height=1.25in,clip,keepaspectratio]{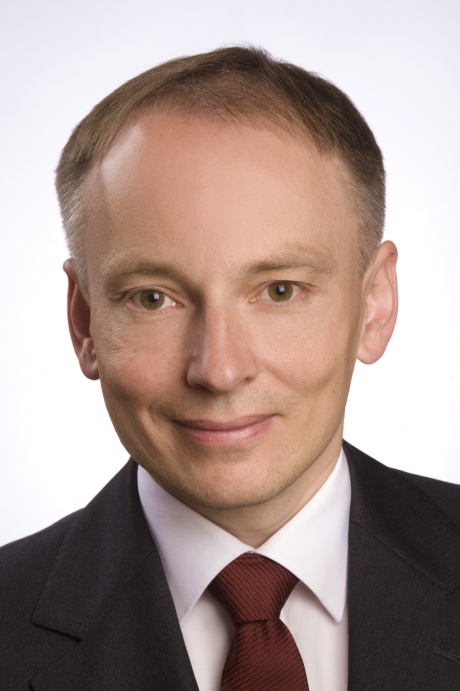}}]{Markus Ulrich} is a distinguished academic and industry expert with over twenty years of experience bridging the realms of academia and industry. Currently a Professor for Machine Vision Metrology at the Institute of Photogrammetry and Remote Sensing at the Karlsruhe Institute of Technology (KIT), his journey began with a PhD degree at the Department of Civil, Geo, and Environmental Engineering of Technical University of Munich (TUM). Prior to academia, he served as the Invention and Patent Manager at MVTec Software GmbH, Munich, and headed the research team at the same institution. In 2017, he completed his habilitation (venia legendi) and was appointed as a Privatdozent (Lecturer) at KIT. His research interests include machine vision, close-range photogrammetry, machine learning, and their applications in industry for quality inspection, automation, and robotics.
\end{IEEEbiography}

\bibliographystyle{IEEEtran.bst}
\bibliography{library}

\begin{thebibliography}{10}
\providecommand{\url}[1]{#1}
\csname url@samestyle\endcsname
\providecommand{\newblock}{\relax}
\providecommand{\bibinfo}[2]{#2}
\providecommand{\BIBentrySTDinterwordspacing}{\spaceskip=0pt\relax}
\providecommand{\BIBentryALTinterwordstretchfactor}{4}
\providecommand{\BIBentryALTinterwordspacing}{\spaceskip=\fontdimen2\font plus
\BIBentryALTinterwordstretchfactor\fontdimen3\font minus \fontdimen4\font\relax}
\providecommand{\BIBforeignlanguage}[2]{{%
\expandafter\ifx\csname l@#1\endcsname\relax
\typeout{** WARNING: IEEEtran.bst: No hyphenation pattern has been}%
\typeout{** loaded for the language `#1'. Using the pattern for}%
\typeout{** the default language instead.}%
\else
\language=\csname l@#1\endcsname
\fi
#2}}
\providecommand{\BIBdecl}{\relax}
\BIBdecl

\bibitem{Uncertainty_decomposition}
E.~Huellermeier and W.~Waegeman, ``{Aleatoric and epistemic uncertainty in machine learning: an introduction to concepts and methods},'' \emph{Machine Learning}, vol. 110, 2021, {URL:} \url{https://doi.org/10.1007/s10994-021-05946-3}.

\bibitem{Mobileye}
I.~B. Shalom, ``Systems and methods for detecting traffic lights,'' U.S. Patent US9\,272\,709B2, March 1, 2016, {URL:} \url{https://patents.google.com/patent/US9272709B2/en}.

\bibitem{Tesla}
J.~Emmons, D.~Hung, E.~Knight, and L.~McIntosh, ``Vision-based machine learning model for autonomous driving with adjustable virtual camera,'' U.S. Patent US20\,230\,057\,509A1, February 23, 2023, {URL:} \url{https://patents.google.com/patent/US20230057509A1/en}.

\bibitem{Nvidia}
Y.~Chen, B.~Ivanovic, and M.~Pavone, ``Neural network trajectory prediction,'' U.S. Patent US20\,230\,391\,374A1, December 7, 2023, {URL:} \url{https://patents.google.com/patent/US20230391374A1/en}.

\bibitem{Wolf_ICCV2023}
D.~W. Wolf, M.~Ulrich, and N.~Kapoor, ``{Sensitivity analysis of AI-based algorithms for autonomous driving on optical wavefront aberrations induced by the windshield},'' \emph{2023 IEEE/CVF International Conference on Computer Vision Workshops (ICCVW)}, pp. 4102--4111, 10 2023, {URL:} \url{https://ieeexplore.ieee.org/document/10350923/}.

\bibitem{mECE_mUCE}
M.~Pakdaman~Naeini, G.~Cooper, and M.~Hauskrecht, ``{Obtaining Well Calibrated Probabilities Using Bayesian Binning},'' \emph{Proceedings of the AAAI Conference on Artificial Intelligence}, vol.~29, no.~1, 2 2015, {URL:} \url{https://ojs.aaai.org/index.php/AAAI/article/view/9602}.

\bibitem{dreissig2023calibration}
M.~Dreissig, F.~Piewak, and J.~Boedecker, ``{On the Calibration of Uncertainty Estimation in LiDAR-based Semantic Segmentation},'' in \emph{{IEEE 26th International Conference on Intelligent Transportation Systems (ITSC)}}.\hskip 1em plus 0.5em minus 0.4em\relax IEEE, 2023, pp. 4798--4805.

\bibitem{Shannon}
C.~E. Shannon, ``{A mathematical theory of communication},'' \emph{The Bell System Technical Journal}, vol.~27, no.~3, pp. 379--423, 1948.

\bibitem{maag2020time}
K.~Maag, M.~Rottmann, and H.~Gottschalk, ``{Time-dynamic estimates of the reliability of deep semantic segmentation networks},'' \emph{IEEE 32nd International Conference on Tools with Artificial Intelligence (ICTAI)}, pp. 502--509, 2020.

\bibitem{Wolf_GCPR24}
D.~W. Wolf, P.~Balaji, A.~Braun, and M.~Ulrich, ``{Decoupling of neural network calibration measures},'' \emph{German Conference on Pattern Recognition (GCPR)}, 2024.

\bibitem{PTS}
C.~Tomani, D.~Cremers, and F.~Buettner, ``Parameterized temperature scaling for boosting the expressive power in post-hoc uncertainty calibration,'' \emph{European Conference on Computer Vision (ECCV)}, pp. 555--569, 2022.

\bibitem{PINNs_CV}
C.~Banerjee, K.~Nguyen, C.~Fookes, and K.~George, ``{Physics-Informed Computer Vision: A Review and Perspectives},'' \emph{ACM Comput. Surv.}, vol.~57, no.~1, Oct. 2024, {URL:} \url{https://doi.org/10.1145/3689037}.

\bibitem{TS}
C.~Guo, G.~Pleiss, Y.~Sun, and K.~Q. Weinberger, ``{On calibration of modern neural networks},'' \emph{International conference on machine learning (ICML)}, pp. 1321--1330, 2017.

\bibitem{States_of_matter_Goodstein}
D.~L. Goodstein, \emph{{States of matter}}, ser. Prentice-Hall physics series.\hskip 1em plus 0.5em minus 0.4em\relax Englewood Cliffs, NJ: Prentice-Hall, 1975.

\bibitem{Calibrating_Language_Models}
J.~Xie, A.~S. Chen, Y.~Lee, E.~Mitchell, and C.~Finn, ``Calibrating language models with adaptive temperature scaling,'' in \emph{Proceedings of the 2024 Conference on Empirical Methods in Natural Language Processing}, Y.~Al-Onaizan, M.~Bansal, and Y.-N. Chen, Eds.\hskip 1em plus 0.5em minus 0.4em\relax Miami, Florida, USA: Association for Computational Linguistics, Nov. 2024, pp. 18\,128--18\,138, {URL:} \url{https://aclanthology.org/2024.emnlp-main.1007}.

\bibitem{ETS}
J.~Zhang, B.~Kailkhura, and T.~Y.-J. Han, ``{Mix-n-Match: ensemble and compositional methods for uncertainty calibration in deep learning},'' in \emph{{Proceedings of the 37th International Conference on Machine Learning}}, ser. ICML'20.\hskip 1em plus 0.5em minus 0.4em\relax JMLR.org, 2020.

\bibitem{Sample_dependent_ATS}
T.~Joy, F.~Pinto, S.-N. Lim, P.~H.~S. Torr, and P.~K. Dokania, ``{Sample-dependent adaptive temperature scaling for improved calibration},'' in \emph{{Proceedings of the Thirty-Seventh AAAI Conference on Artificial Intelligence and Thirty-Fifth Conference on Innovative Applications of Artificial Intelligence and Thirteenth Symposium on Educational Advances in Artificial Intelligence}}, ser. AAAI'23/IAAI'23/EAAI'23.\hskip 1em plus 0.5em minus 0.4em\relax AAAI Press, 2023, {URL:} \url{https://doi.org/10.1609/aaai.v37i12.26742}.

\bibitem{Variational_Autoencoders}
D.~P. Kingma and M.~Welling, ``{Auto-Encoding Variational Bayes},'' \emph{CoRR}, vol. abs/1312.6114, 2013, {URL:} \url{https://api.semanticscholar.org/CorpusID:216078090}.

\bibitem{ATS}
G.~Krumpl, H.~Avenhaus, H.~Possegger, and H.~Bischof, ``{ATS: Adaptive Temperature Scaling for Enhancing Out-of-Distribution Detection Methods},'' in \emph{{2024 IEEE/CVF Winter Conference on Applications of Computer Vision (WACV)}}, 2024, pp. 3852--3861.

\bibitem{p_values}
\BIBentryALTinterwordspacing
A.~Rudolph, J.~Krois, and K.~Hartmann. (2023) {Statistics and Geodata Analysis using Python (SOGA-Py), Department of Earth Sciences, Freie Universitaet Berlin}. [Online]. Available: \url{https://www.geo.fu-berlin.de/en/v/soga-py/Basics-of-statistics/Hypothesis-Tests/Introduction-to-Hypothesis-Testing/Critical-Value-and-the-p-Value-Approach/index.html}
\BIBentrySTDinterwordspacing

\bibitem{Oxford_Dictionary_of_Statistics}
G.~Upton and I.~Cook, \emph{{A Dictionary of Statistics}}, ser. Oxford Paperback Reference.\hskip 1em plus 0.5em minus 0.4em\relax OUP Oxford, 2008, {URL:} \url{https://books.google.de/books?id=u97pzxRjaCQC}.

\bibitem{survey_uncertainty_DNN}
J.~Gawlikowski, C.~R.~N. Tassi, M.~Ali, J.~Lee, M.~Humt, J.~Feng, A.~Kruspe, R.~Triebel, P.~Jung, R.~Roscher, M.~Shahzad, W.~Yang, R.~Bamler, and X.~X. Zhu, ``A survey of uncertainty in deep neural networks,'' \emph{{Artificial Intelligence Review}}, vol.~56, no.~1, pp. 1513--1589, 2023, {URL:} \url{https://doi.org/10.1007/s10462-023-10562-9}.

\bibitem{Deep_ensembles}
B.~Lakshminarayanan, A.~Pritzel, and C.~Blundell, ``{Simple and Scalable Predictive Uncertainty Estimation using Deep Ensembles},'' \emph{Neural Information Processing Systems}, 2016, {URL:} \url{https://api.semanticscholar.org/CorpusID:6294674}.

\bibitem{DUDES}
S.~Landgraf, K.~Wursthorn, M.~Hillemann, and M.~Ulrich, ``{DUDES: Deep Uncertainty Distillation using Ensembles for Semantic Segmentation},'' \emph{PFG --Journal of Photogrammetry, Remote Sensing and Geoinformation Science}, vol.~92, no.~2, pp. 101--114, 2024, {URL:} \url{https://doi.org/10.1007/s41064-024-00280-4}.

\bibitem{EMUFormer}
S.~Landgraf, M.~Hillemann, T.~Kapler, and M.~Ulrich, ``{Efficient multi-task uncertainties for joint semantic segmentation and monocular depth estimation},'' \emph{{German Conference on Pattern Recognition (GCPR)}}, 2024.

\bibitem{Zernike_I}
F.~Zernike and F.~Stratton, ``{Diffraction Theory of the Knife-Edge Test and its Improved Form, The Phase-Contrast Method},'' \emph{Monthly Notices of the Royal Astronomical Society}, vol.~94, no.~5, pp. 377--384, 03 1934, {URL:} \url{https://doi.org/10.1093/mnras/94.5.377}.

\bibitem{Zernike_II}
F.~Zernike, ``{Beugungstheorie des schneidenver-fahrens und seiner verbesserten form, der phasenkontrastmethode},'' \emph{Physica}, vol.~1, no.~7, pp. 689--704, 1934, {URL:} \url{https://www.sciencedirect.com/science/article/pii/S0031891434802595}.

\bibitem{Zernike_III}
A.~B. Bhatia and E.~Wolf, ``{On the circle polynomials of Zernike and related orthogonal sets},'' \emph{Mathematical Proceedings of the Cambridge Philosophical Society}, vol.~50, no.~1, pp. 40--48, 1954.

\bibitem{Astronomy_Zernike_idea}
P.~C. McGuire, D.~G. Sandler, M.~Lloyd-Hart, and T.~A. Rhoadarmer, ``{Adaptive optics: Neural network wavefront sensing, reconstruction, and prediction},'' in \emph{{Scientific Applications of Neural Nets}}, J.~W. Clark, T.~Lindenau, and M.~L. Ristig, Eds.\hskip 1em plus 0.5em minus 0.4em\relax Berlin, Heidelberg: Springer Berlin Heidelberg, 1999, pp. 97--138.

\bibitem{Astronomy_Zernike_ResNet}
T.~E. Andersen, M.~Owner-Petersen, and A.~Enmark, ``{Image-based wavefront sensing for astronomy using neural networks},'' \emph{Journal of Astronomical Telescopes, Instruments, and Systems}, vol.~6, no.~3, p. 034002, 2020, {URL:} \url{https://doi.org/10.1117/1.JATIS.6.3.034002}.

\bibitem{adaptive_optics_EOS}
F.~Merkle and N.~Hubin, ``{Adaptive Optics for the ESO Very Large Telescope},'' in \emph{{Adaptive Optics for Large Telescopes}}.\hskip 1em plus 0.5em minus 0.4em\relax Optica Publishing Group, 1992, {URL:} \url{https://opg.optica.org/abstract.cfm?URI=AOLT-1992-ATuA4}.

\bibitem{adaptive_optics}
K.~M. Hampson, R.~Turcotte, D.~T. Miller, K.~Kurokawa, J.~R. Males, N.~Ji, and M.~J. Booth, ``{Adaptive optics for high-resolution imaging},'' \emph{Nature Reviews Methods Primers}, vol.~1, no.~1, p.~68, 2021, {URL:} \url{https://doi.org/10.1038/s43586-021-00066-7}.

\bibitem{Physics_Driven_Restoration}
A.~Jaiswal, X.~Zhang, S.~H. Chan, and Z.~Wang, ``{Physics-Driven Turbulence Image Restoration with Stochastic Refinement},'' in \emph{{2023 IEEE/CVF International Conference on Computer Vision (ICCV)}}.\hskip 1em plus 0.5em minus 0.4em\relax Los Alamitos, CA, USA: IEEE Computer Society, Oct 2023, pp. 12\,136--12\,147, {URL:} \url{https://doi.ieeecomputersociety.org/10.1109/ICCV51070.2023.01118}.

\bibitem{vision_transformer}
A.~Dosovitskiy, L.~Beyer, A.~Kolesnikov, D.~Weissenborn, X.~Zhai, T.~Unterthiner, M.~Dehghani, M.~Minderer, G.~Heigold, S.~Gelly, J.~Uszkoreit, and N.~Houlsby, ``{An Image is Worth 16x16 Words: Transformers for Image Recognition at Scale},'' in \emph{{9th International Conference on Learning Representations, {ICLR} 2021, Virtual Event, Austria, May 3-7, 2021}}, 2021.

\bibitem{Chatterjee_correlation}
S.~Chatterjee, ``{A New Coefficient of Correlation},'' \emph{Journal of the American Statistical Association}, vol. 116, no. 536, pp. 2009--2022, 2021.

\bibitem{Dette-Siburg-Stoimenov_correlation}
H.~Shi, M.~Drton, and F.~Han, ``{On the power of Chatterjees rank correlation},'' \emph{Biometrika}, vol. 109, no.~2, pp. 317--333, 10 2021.

\bibitem{Wolf_ITSC2023}
D.~W. Wolf, M.~Ulrich, and A.~Braun, ``{Windscreen Optical Quality for AI Algorithms: Refractive Power and MTF not Sufficient},'' \emph{2023 IEEE 26th International Conference on Intelligent Transportation Systems (ITSC)}, pp. 5190--5197, 9 2023, {URL:} \url{https://ieeexplore.ieee.org/document/10421970/}.

\bibitem{Fourier_optics}
J.~W. Goodman, \emph{{Introduction to Fourier optics}}.\hskip 1em plus 0.5em minus 0.4em\relax McGraw-Hill, 1968.

\bibitem{Control_Systems}
M.~C. Khoo, ``{Physiological Control Systems : Analysis, Simulation, and Estimation},'' 2018, iEEE Press series in biomedical engineering, ISBN: 978-1-119-05879-3.

\bibitem{Green_as_PDF}
M.~Bakker, K.~Maas, and J.~R. Von~Asmuth, ``{Calibration of transient groundwater models using time series analysis and moment matching},'' 2008, water Resources Research, Vol.44. {URL:} \url{https://doi.org/10.1029/2007WR006239}.

\bibitem{MTF_requirement}
P.~Mueller and A.~Braun, ``{MTF as a performance indicator for AI algorithms?}'' \emph{Electronic Imaging}, vol.~35, no.~16, pp. 125--1--125--1, 2023, {URL:} \url{https://library.imaging.org/ei/articles/35/16/AVM-125}.

\bibitem{Plancherel_theorem}
A.~Deitmar and S.~Echterhoff, ``{Principles of Harmonic Analysis},'' 2008, springer New York. ISBN: 9780387854687.

\bibitem{Survey_paper}
S.~Minaee, Y.~Boykov, F.~Porikli, A.~Plaza, N.~Kehtarnavaz, and D.~Terzopoulos, ``{Image Segmentation Using Deep Learning: A Survey},'' 2020, {URL:} \url{https://arxiv.org/abs/2001.05566}.

\bibitem{UCS_Kira}
K.~Wursthorn, M.~Hillemann, and M.~Ulrich, ``{Uncertainty Quantification with Deep Ensembles for 6D Object Pose Estimation},'' \emph{ISPRS Annals of the Photogrammetry, Remote Sensing and Spatial Information Sciences}, vol. X-2-2024, pp. 223--230, 2024, {URL:} \url{https://isprs-annals.copernicus.org/articles/X-2-2024/223/2024/}.

\bibitem{variance_decomposition}
N.~Weiss, P.~Holmes, and M.~Hardy, \emph{{A Course in Probability}}.\hskip 1em plus 0.5em minus 0.4em\relax Pearson Addison Wesley, 2005.

\bibitem{Tilt_vs_Blur}
S.~H. Chan, ``{Tilt-Then-Blur or Blur-Then-Tilt? Clarifying the Atmospheric Turbulence Model},'' \emph{IEEE Signal Processing Letters}, vol.~29, pp. 1833--1837, 2022.

\bibitem{Wolf_Metrologia}
D.~W. Wolf, M.~Ulrich, and A.~Braun, ``{Novel developments of refractive power measurement techniques in the automotive world},'' \emph{Metrologia}, vol.~60, 9 2023, {URL:} \url{https://iopscience.iop.org/article/10.1088/1681-7575/acf1a4}.

\bibitem{ResNet}
K.~He, X.~Zhang, S.~Ren, and J.~Sun, ``{Deep Residual Learning for Image Recognition},'' \emph{IEEE Conference on Computer Vision and Pattern Recognition (CVPR)}, pp. 770--778, 2016.

\bibitem{ResNet_VGP}
A.~Veit, M.~Wilber, and S.~Belongie, ``{Residual networks behave like ensembles of relatively shallow networks},'' \emph{Proceedings of the 30th International Conference on Neural Information Processing Systems}, pp. 550--558, 2016.

\bibitem{ResNet_VGP_norm}
A.~Zaeemzadeh, N.~Rahnavard, and M.~Shah, ``{Norm-Preservation: Why Residual Networks Can Become Extremely Deep?}'' \emph{IEEE Transactions on Pattern Analysis and Machine Intelligence}, vol.~43, pp. 3980--3990, 2018, {URL:} \url{https://api.semanticscholar.org/CorpusID:29155803}.

\bibitem{A2D2}
J.~Geyer, Y.~Kassahun, M.~Mahmudi, X.~Ricou, R.~Durgesh, A.~S. Chung, L.~Hauswald, V.~H. Pham, M.~M{\"u}hlegg, S.~Dorn, T.~Fernandez, M.~J{\"a}nicke, S.~Mirashi, C.~Savani, M.~Sturm, O.~Vorobiov, M.~Oelker, S.~Garreis, and P.~Schuberth, ``{A2D2: Audi Autonomous Driving Dataset},'' \emph{arXiv}, 2020, {URL:} \url{https://www.a2d2.audi}.

\bibitem{Cordts2016Cityscapes}
M.~Cordts, M.~Omran, S.~Ramos, T.~Rehfeld, M.~Enzweiler, R.~Benenson, U.~Franke, S.~Roth, and B.~Schiele, ``{The Cityscapes Dataset for Semantic Urban Scene Understanding},'' in \emph{{Proc. of the IEEE Conference on Computer Vision and Pattern Recognition (CVPR)}}, 2016.

\bibitem{Focal_loss}
T.-Y. Lin, P.~Goyal, R.~Girshick, K.~He, and P.~Dollar, ``{Focal Loss for Dense Object Detection},'' \emph{IEEE International Conference on Computer Vision (ICCV)}, pp. 2999--3007, 2017, {URL:} \url{https://doi.ieeecomputersociety.org/10.1109/ICCV.2017.324}.

\bibitem{L1_over_L2}
H.~Zhao, O.~Gallo, I.~Frosio, and J.~Kautz, ``{Loss Functions for Image Restoration With Neural Networks},'' \emph{IEEE Transactions on Computational Imaging}, vol.~3, no.~1, pp. 47--57, 2017.

\bibitem{Large_batch_sizes}
N.~S. Keskar, D.~Mudigere, J.~Nocedal, M.~Smelyanskiy, and P.~T.~P. Tang, ``{On Large-Batch Training for Deep Learning: Generalization Gap and Sharp Minima},'' \emph{5th International Conference on Learning Representations, {ICLR} 2017, Toulon, France, April 24-26, 2017, Conference Track Proceedings}, 2017.

\bibitem{GELU}
D.~Hendrycks and K.~Gimpel, ``{Gaussian Error Linear Units (GELUs)},'' \emph{arXiv}, 2023, {URL:} \url{https://arxiv.org/abs/1606.08415}.

\bibitem{RELU}
V.~Nair and G.~E. Hinton, ``{Rectified linear units improve restricted Boltzmann machines},'' \emph{Proceedings of the 27th International Conference on International Conference on Machine Learning (ICML)}, pp. 807--814, 2010.

\bibitem{Non_linear_activation_function}
C.~Bauckhage and D.~Speicher, ``{Lecture Notes on Machine Learning: Neurons with Non-Monotonic Activation Functions},'' University of Bonn, Tech. Rep., 07 2019.

\bibitem{ELU}
D.~Clevert, T.~Unterthiner, and S.~Hochreiter, ``Fast and accurate deep network learning by exponential linear units (elus),'' in \emph{4th International Conference on Learning Representations, {ICLR} 2016, San Juan, Puerto Rico, May 2-4, 2016, Conference Track Proceedings}, 2016, {URL:} \url{http://arxiv.org/abs/1511.07289}.

\bibitem{KOR}
T.~Kim and S.-Y. Yun, ``{Revisiting Orthogonality Regularization: A Study for Convolutional Neural Networks in Image Classification},'' \emph{IEEE Access}, vol.~10, pp. 69\,741--69\,749, 2022.

\bibitem{Deep_optics_I}
J.~Chang and G.~Wetzstein, ``{Deep Optics for Monocular Depth Estimation and 3D Object Detection},'' in \emph{{2019 IEEE/CVF International Conference on Computer Vision (ICCV)}}, 2019, pp. 10\,192--10\,201.

\bibitem{Deep_optics_II}
E.~Tseng, A.~Mosleh, F.~Mannan, K.~St-Arnaud, A.~Sharma, Y.~Peng, A.~Braun, D.~Nowrouzezahrai, J.-F. Lalonde, and F.~Heide, ``{Differentiable Compound Optics and Processing Pipeline Optimization for End-to-end Camera Design},'' \emph{ACM Transactions on Graphics}, vol.~40, no.~2, 6 2021, {URL:} \url{https://doi.org/10.1145/3446791}.

\bibitem{Deep_optics_III}
X.~Yang, Q.~Fu, Y.~Nie, and W.~Heidrich, ``{Image Quality Is Not All You Want: Task-Driven Lens Design for Image Classification},'' 05 2023, {URL:} \url{https://doi.org/10.48550/arXiv.2305.17185}.

\bibitem{GUM_compliance}
B.~Ludwig, ``{GUM-compliant neural network robustness verification},'' \emph{Master thesis at the Technical University of Berlin, Zuse Institute Berlin and Physikalisch-Technische Bundesanstalt}, 2023.

\bibitem{Monte_Carlo}
S.~Raychaudhuri, ``{Introduction to Monte Carlo simulation},'' \emph{Winter Simulation Conference}, pp. 91--100, 2008.

\bibitem{GUM}
J.~C. for Guides~in Metrology~(JCGM), ``{Evaluation of measurement data, Guide to the expression of uncertainty in measurement (GUM)},'' \emph{International Bureau of Weights and Measures (BIPM)}, 2008, {URL:} \url{https://www.bipm.org/documents/20126/2071204/JCGM_100_2008_E.pdf/cb0ef43f-baa5-11cf-3f85-4dcd86f77bd6}.

\bibitem{GUM_II}
B.~Pesch, \emph{{Bestimmung der Messunsicherheit nach GUM, Grundlagen der Metrologie}}.\hskip 1em plus 0.5em minus 0.4em\relax Books on Demand (BoD), 2003.

\bibitem{Ridge_regression_I}
\BIBentryALTinterwordspacing
M.~Taboga. (2021) {Ridge regression, Lectures on probability theory and mathematical statistics. Kindle Direct Publishing}. [Online]. Available: \url{https://www.statlect.com/fundamentals-of-statistics/ridge-regression}
\BIBentrySTDinterwordspacing

\bibitem{Ridge_regression_II}
C.~M. Theobald, ``{Generalizations of Mean Square Error Applied to Ridge Regression},'' \emph{Journal of the Royal Statistical Society. Series B (Methodological)}, vol.~36, no.~1, pp. 103--106, 1974, {URL:} \url{http://www.jstor.org/stable/2984775}.

\bibitem{Ridge_regression_III}
R.~W. Farebrother, ``Further results on the mean square error of ridge regression,'' \emph{Journal of the Royal Statistical Society. Series B (Methodological)}, vol.~38, no.~3, pp. 248--250, 1976, {URL:} \url{http://www.jstor.org/stable/2984971}.

\bibitem{Gaussian_NLL}
D.~Nix and A.~Weigend, ``Estimating the mean and variance of the target probability distribution,'' in \emph{Proceedings of 1994 IEEE International Conference on Neural Networks (ICNN'94)}, vol.~1, no.~1, 1994, pp. 55--60.

\bibitem{Cramer}
H.~Cram\'er, \emph{Mathematical Methods of Statistics}.\hskip 1em plus 0.5em minus 0.4em\relax Princeton: Princeton University Press, 1946, {URL:} \url{https://doi.org/10.1515/9781400883868}.

\bibitem{Rao}
C.~Rao, ``{Information and the Accuracy Attainable in the Estimation of Statistical Parameters},'' \emph{{Bulletin of Calcutta Mathematical Society}}, vol.~37, pp. 81--91, 1945.

\bibitem{Frechet}
M.~Fr\'echet, ``{Sur l'extension de certaines evaluations statistiques au cas de petits echantillons},'' \emph{{Revue de l'Institut International de Statistique}}, vol.~11, no.~3, pp. 182--205, 1943, {URL:} \url{http://www.jstor.org/stable/1401114}.

\bibitem{Braun_TM}
A.~Braun, ``{Automotive mass production of camera systems: Linking image quality to AI performance},'' \emph{tm - Technisches Messen}, vol.~90, no.~3, pp. 205--218, 2023, {URL:} \url{https://doi.org/10.1515/teme-2022-0029}.

\bibitem{Vee_model}
D.~D. Walden, T.~M. Shortell, G.~J. Roedler, B.~A. Delicado, O.~Mornas, Y.~Yew-Seng, and D.~Endler, \emph{INCOSE Systems Engineering Handbook}, 5th~ed.\hskip 1em plus 0.5em minus 0.4em\relax Wiley, 6 2023, {URL:} \url{https://www.wiley.com/en-ie/INCOSE+Systems+Engineering+Handbook%2C+5th+Edition-p-9781119814313}.

\bibitem{Wolf_IMEKO2024}
D.~W. Wolf, B.~Thielbeer, M.~Ulrich, and A.~Braun, ``{Wavefront aberration measurements based on the Background Oriented Schlieren method},'' \emph{Measurement: Sensors}, p. 101509, 2024, {URL:} \url{https://doi.org/10.1016/j.measen.2024.101509}.

\end{thebibliography}
\end{document}